\documentclass{article}

\usepackage[nonatbib,main,final]{neurips_2026}
\usepackage[url=false, doi=false, isbn=false,natbib=true,eprint=false]{biblatex}
\usepackage{amsmath}
\usepackage{amsthm}
\usepackage{amssymb}
\DeclareMathOperator{\E}{\mathbb{E}}
\usepackage[createShortEnv]{proof-at-the-end}

\usepackage[utf8]{inputenc} 
\usepackage[T1]{fontenc}    
\usepackage{hyperref}       
\usepackage{cleveref}       
\theoremstyle{plain}

\theoremstyle{definition}

\usepackage{url}            
\usepackage{booktabs}       
\usepackage{amsfonts}       
\usepackage{nicefrac}       
\usepackage{microtype}      
\usepackage{wrapfig}
\usepackage{xcolor}         
\usepackage[inkscapelatex=false,inkscapearea=drawing]{svg}
\usepackage{multirow}
\usepackage{algorithm}
\usepackage[noend]{algpseudocode}
\usepackage{algorithm}
\usepackage{algpseudocode}
\usepackage{bm}
\usepackage{enumitem}
\newcommand{\method}{\textsc{SFTProbe}}

\title{From Weeks to Hours: Fast and Principled\\ SFT Data Generation for Reasoning LLMs}

\author{
Hongyi Henry Jin$^\dagger$\qquad Wenhan Yang$^\dagger$\\
\textbf{Meysam Ghaffari$^\star$\qquad Carlos Morato$^\star$\qquad Baharan Mirzasoleiman$^\dagger$ }\\
$^\dagger$University of California, Los Angeles \qquad $^\star$Optum AI\\
\texttt{\{hongyi,wenhan\_yang,baharan\}@cs.ucla.edu}
}

\begin{document}

\maketitle

\begin{abstract}
Reasoning-oriented post-training enables large language models (LLMs) to solve complex tasks via multi-step inference. Supervised fine-tuning (SFT) on high-quality reasoning traces is a particularly efficient approach, but its effectiveness depends critically on careful data curation. Existing pipelines rely on a costly generate-then-filter paradigm: large pools of long reasoning traces are produced by multiple teacher models and subsequently filtered for difficulty and diversity using additional LLMs, often requiring weeks of computation. In this work, we propose \method, a fundamentally different approach that bypasses this pipeline by \textit{directly predicting} the quality of reasoning data. 
We show that a brief LoRA-based adaptation, combined with evaluating loss on only 5\% of each reasoning trace, suffices to estimate both difficulty and diversity. \method\  is grounded in a geometric view of post-training, where pretrained LLMs lie in a low-rank, anisotropic loss basin. By probing this basin via low-rank perturbations, we construct compact loss signatures whose clustering captures diversity, while loss at curvature inflection points provides a robust proxy for difficulty. We also provide a simple criterion for teacher selection. Empirically, our approach reduces data curation time from weeks to hours while maintaining or improving performance. Remarkably, on mathematical reasoning, 5k generated examples match the highly filtered 89k-example in OpenThoughts-Math dataset, and 7.5k generated examples outperforms it by 2.5\% in accuracy. We further introduce a high-quality physics reasoning dataset, and use it to train a model that outperforms an IPhO 2025 silver-medal-winning model. The code and the dataset are released at \url{https://bigml-cs-ucla.github.io/SFTProbe-page/}. \looseness=-1
\end{abstract}
\section{Introduction}
Large language models (LLMs) have recently demonstrated an emerging ability to solve complex tasks by allocating substantial computation to \textit{reasoning at inference time} \cite{OpenAI24,guo2025deepseek,s1-tts,M23k}. Unlike instruction tuning—where models learn to produce answers in a single pass, akin to fast “System 1” thinking—reasoning fine-tuning encourages slower, deliberative “System 2” behavior, where models generate intermediate steps and refine them iteratively. Enabling such capabilities relies on post-training, either through large-scale reinforcement learning (RL) \cite{guo2025deepseek,OpenAI24} or supervised fine-tuning (SFT) on high-quality reasoning traces \cite{s1-tts,M23k,openthoughts}. While RL is effective but costly, SFT is far more efficient, yet highly sensitive to data quality—particularly the diversity and difficulty of reasoning trajectories \cite{javanmard2026theoretical,javanmardunderstanding}—making data curation a key bottleneck.

Reasoning data differs fundamentally from standard instruction-tuning corpora. Instead of short question–answer pairs, reasoning traces can span thousands—or even hundreds of thousands—of tokens \cite{s1-tts,openthoughts}. Current pipelines therefore rely on a costly generate-then-filter paradigm: multiple strong teacher models are used to produce large pools of candidate traces, which are subsequently filtered using LLM-based heuristics for difficulty and diversity, followed by downstream training to evaluate their utility \cite{M23k,openthoughts}. This process is not only computationally expensive but also slow, often requiring weeks of iterative generation, filtering, and evaluation. These limitations raise a fundamental question: 
\begin{wrapfigure}{r}{0.42\textwidth}
    \centering
    \vspace{-1mm}
    \includesvg[width=\linewidth]{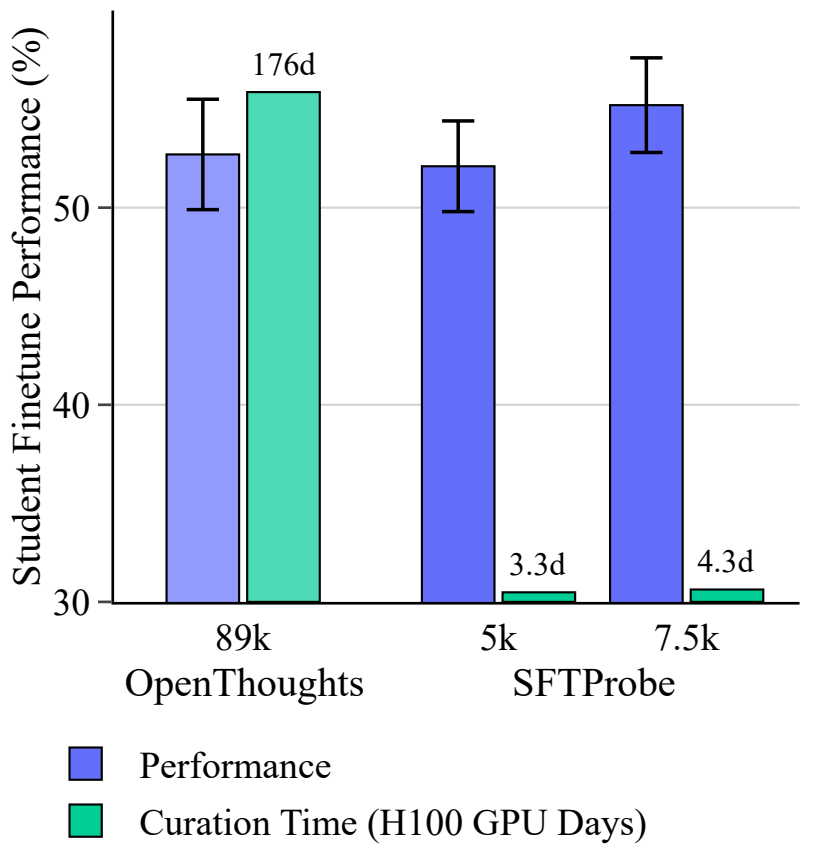}
    \vspace{-8mm}
    \caption{Comparison of the curated math reasoning dataset in OpenThoughts \citep{openthoughts} and \method. Training on 5k curated data matches the performance with 53x speedup, and on 7.5k improves the performance by 2.5\% with 41x speedup.}
    \label{fig:intro-numina}
    \vspace{-5mm}
\end{wrapfigure}
\textit{can we directly generate high-quality reasoning data without resorting to large-scale sampling and filtering?} 
This is very challenging because the quality of a reasoning trace—particularly its difficulty and diversity—is typically only observable after full generation and training.\looseness=-1

In this work, we introduce a principled and efficient alternative that bypasses this pipeline entirely.\method\  builds on a recent observation that the loss over the initial portion of reasoning traces, evaluated along the training trajectory, provides useful signals about the quality of full reasoning \cite{ICML}. However, prior methods require training on a large corpus of full reasoning traces to obtain such trajectories. 
We relax this requirement and show that a brief LoRA-based adaptation on a small set of fully generated examples, is sufficient to reliably estimate both difficulty and diversity. This enables the direct construction of high-quality SFT datasets without exhaustive generation or filtering, reducing data curation time from weeks to minutes.

\method\  is grounded in a geometric perspective of post-training, in which pretrained LLMs are understood to operate within a low-rank loss basin characterized by relatively flat curvature \cite{fort2020deep}. From a theoretical standpoint, we show that examples with similar loss values at the {endpoints} of this basin—corresponding to \textit{inflection points} where curvature changes—exhibit tightly bounded differences in their gradients during SFT, and therefore behave similarly under training. However, the loss landscape is inherently anisotropic: curvature varies significantly across directions, implying that reaching the basin boundary requires direction-dependent perturbation magnitudes.

To account for this heterogeneity, we perturb the model along the LoRA directions to reach the boundary of the basin, and thereby obtain a set of representative checkpoints. We then evaluate the loss on only the initial portion of each reasoning trace (as little as 5\% of tokens) across these checkpoints, producing a compact \textit{loss signature} for every example. Clustering these signatures provably groups examples with bounded gradient variation during SFT, providing a principled notion of diversity without requiring full training. In addition, we identify a highly reliable proxy for difficulty based on the loss at the inflection boundary of the basin, again computable from only a tiny prefix of the reasoning trajectory. Finally, this geometric framework yields a simple and efficient criterion for teacher selection: the best teacher is the one that minimizes early-token loss under the pretrained model, aligning data quality with the intrinsic structure of the loss landscape.

We validate \method\  on medical, mathematical, and physical reasoning tasks. Remarkably, with only 5K directly generated examples, \method\  matches the performance of large-scale datasets such as OpenThoughts, which contains 89K carefully filtered examples constructed through expensive pipelines; and with 7.5k generated examples, out approach outperforms it by 2.5\% in accuracy. In the domain of physics, we further present, to our knowledge, the first high-quality SFT dataset for reasoning. For medical reasoning, \method\  outperforms data selection baselines, with substantially lower costs.
Together, these results demonstrate that high-quality reasoning data can be \textit{predicted rather than exhaustively searched}, offering a new direction for scalable and principled post-training of LLMs. 
\section{Related works}

\subsection{Data Curation for Reasoning Models}

Supervised fine-tuning (SFT) is highly sensitive to the quality—i.e. difficulty and diversity—\cite{javanmard2026theoretical} of reasoning traces, and simply increasing the amount of data without proper curation can even degrade performance \cite{akter2025front}. 
Unlike instruction tuning data which consists of short (instruction, output) pairs, reasoning trajectories are highly exploratory and noisy: Throughout the long chain-of-thought, the model frequently makes multiple failed attempts, pauses to re-evaluate (e.g., generating “Wait, but the problem says…” or “Hmm, let me double check…”), and tries multiple ways to approach the problem before arriving at a solution (c.f. \Cref{fig:reasoning-noisy}). This inherent noise disrupts data selection metrics that are applicable to instruction tuning, including difficulty-filtering based on
perplexity \cite{IFD-selection,marion2023less}, learnability-based (loss reduction during training) \cite{DavIR-learnability-loss}, and gradient magnitude \cite{GraDS}, and diversity-filtering based on embedding similarity \cite{instruction-tuning-selection,bhatt2024experimental} and the superior gradient similarity \cite{TAGCOS,S2L}.

Due to these limitations, reasoning traces are often either manually curated \cite{s1-tts} or generated by strong teacher models (e.g., DeepSeek-R1), and then filtered by two key properties: diversity and difficulty \cite{openthoughts,s1-tts,M23k}, meaning they include a broad range of examples that are sufficiently challenging for the target model. In practice, this filtering relies heavily on LLM-based heuristics. Diversity is encouraged by categorizing and selecting examples across different groups, while difficulty is commonly enforced by prompting a strong model to score question hardness and keeping only the most challenging ones, which outperforms the reasoning length heuristic. Despite their effectiveness, these approaches are computationally expensive: generating large numbers of long reasoning traces with strong models, followed by extensive filtering, makes the construction of high-quality SFT datasets costly and difficult to scale. 
Recent study of \citet{ICML} observed that the loss over the initial portion (100-1K tokens) of reasoning traces, evaluated along the fine-tuning trajectory, provides useful signals about the quality of full reasoning. However, this requires training on a large corpus of \textit{full} reasoning traces to obtain the fine-tuning trajectory and does not enable efficient SFT data curation.\looseness=-1

In our work, we develop a principled method to directly \textit{predict} the quality of reasoning without having to fully generate them and without relying on accurate fine-tuning trajectories.

\subsection{Teacher Selection for LLMs}
Selecting appropriate teacher models commonly prioritizes alignment with student capabilities. 
For example, GRACE \cite{panigrahi2025good} leverages a cross-validation structure, where it takes gradients from one split of data and measures their magnitudes re-weighted under the spectrum of the other split. The gradient spectrum reflects data diversity, while the gradient norm encodes teacher-student alignment. 
Most recently, \citep{teacher-selection} introduced Local Naturalness, which scores a response by measuring the student’s log-probabilities over short, sequential reasoning steps (e.g., sentences) conditioned only on a small local window, and used aggregated
local scores across prompts for teacher selection. These methods, however, require generating full reasoning traces for all the examples for every teacher model, and hence incur substantial computational cost.

In our work, we show that the best teacher model can be efficiently selected as the one that minimizes average early-token losses under the pretrained model.
\section{Efficient SFT Data Curation}
In this section, we discuss generating high-quality—i.e., diverse and difficult—SFT dataset, given a teacher model, without having to fully generate reasoning traces and training on them. 

\begin{figure}[t]
    \centering
    \includesvg[width=\linewidth]{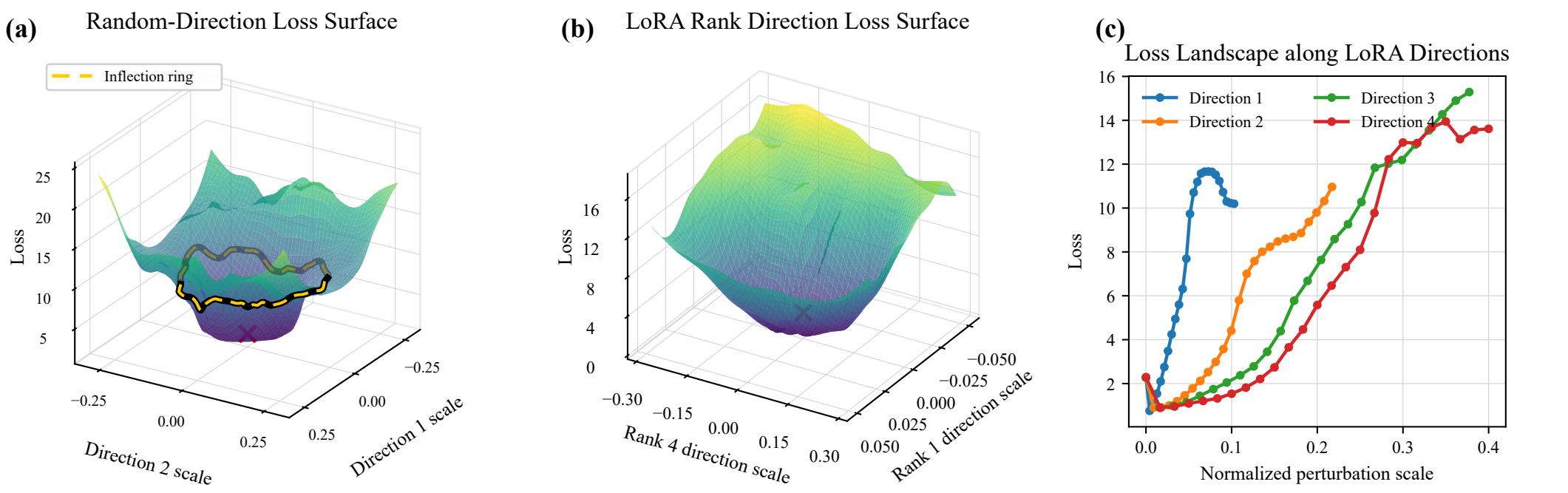}
    \vspace{-5mm}
    \caption{(a) Loss landscape of Qwen2.5-3B-Instruct along two random directions. (b) LoRA directions have different curvatures. (c) Directions with lower curvature should be perturbed more to reach the inflection point.}
    \label{fig:loss-3d}
    \vspace{-3mm}
\end{figure}

\subsection{Predicting Similarity of Reasoning Traces}
\textbf{Gradient similarity.} During training, each example updates the model parameters through its gradient. Consequently, examples with similar gradients induce similar parameter updates and therefore provide largely redundant training signals \cite{mirzasoleiman2020coresets,nguyenmini}. Motivated by this observation, prior works \cite{S2L,ICML} estimate data diversity by clustering examples according to their gradient similarity across \textit{multiple checkpoints along the fine-tuning trajectory}. While effective, this approach requires fully training the model on fully generated reasoning traces in order to obtain the trajectory itself, making it computationally expensive. Moreover, because the training trajectory is inherently stochastic—due to mini-batch ordering, optimizer noise, and other sources of randomness—the resulting gradient similarities can be noisy and unstable. Here, we develop a significantly more efficient and reliable approach for predicting gradient similarity without having to fully generate reasoning traces and training on them.\looseness=-1

\textbf{Fine-tuning loss.} Our key observation is that fine-tuning occurs within a \textit{low-rank}, relatively \textit{flat}, and smooth region of the loss landscape, where the loss around the pretrained initialization is well approximated by a \textit{near-parabolic} geometry \citep{fort2020deep}. \Cref{fig:loss-3d} (a) illustrates the loss landscape of Qwen2.5-3B-Instruct around the pretrained checkpoint. In this regime, the geometry of the loss basin provides substantially richer information than the stochastic training trajectory itself. Specifically, we show that loss difference at the pretrained parameters and \textit{boundaries} of the parabolic basin—corresponding to inflection points where the curvature changes (shown as the yellow band in \Cref{fig:loss-3d}(a))—induces the tightest bound on gradient similarity throughout fine-tuning.  Intuitively, if two examples behave similarly at the extremities of the basin, then their optimization dynamics remain similar across the entire training process. Importantly, bounded loss differences at the pretrained checkpoint and the inflection points yield significantly tighter estimates of gradient similarity than measuring loss similarity along noisy training trajectories. This shifts the problem from tracking optimization dynamics to accurately characterizing the local geometry of the loss basin, as we discuss next.\looseness=-1

\subsubsection{Approximating the Near-Parabolic Low-rank Loss Basin}\label{sec:parabolic}\label{sec:parabolic}

Since fine-tuning happens in a low-rank region, only a small number of parameter directions dominate learning. We exploit this structure by briefly training a LoRA adapter on a very small subset of reasoning traces to identify these dominant directions. Importantly, this short adaptation phase is not intended to meaningfully optimize the model itself; rather, it serves only to estimate the principal directions of the local loss geometry. Thus, very brief training on a small subset is sufficient.

However, the loss landscape is anisotropic: as shown in \Cref{fig:loss-3d}(b), curvature varies substantially across directions, implying that the extent of the near-parabolic region differs from one direction to another. To accurately estimate the boundary of the basin, we therefore perturb the model incrementally along each direction, progressively increasing the perturbation magnitude while the loss continues to rise smoothly, and stopping once this behavior breaks down (\textit{c.f.} \Cref{fig:loss-3d}(c)). We then fit a cubic spline to the resulting loss profile and identify the inflection point where the curvature changes sign. Repeating this procedure across all LoRA directions provides a fast and reliable approximation of the inflection boundaries of the near-parabolic loss basin. We provide the pseudocode for finding the inflection point in \Cref{alg:inflection-search}. 

\begin{algorithm}[t]
\caption{Inflection point finder. Detailed implementation in \Cref{alg:inflection-search}}
\label{alg:high-level-lora}
\begin{algorithmic}[1]
\Require Perturbation direction $v$

    \State Incrementally perturb base model $\theta_0$ while $\E_{x,y}\mathcal{L}_{}(x,y,\theta_0+\epsilon\cdot v)$ is monotone
    \State Fit a cubic spline surrogate $f(\epsilon)$ on $\left\{\left(\epsilon, \E_{x,y}\mathcal{L}_{}(x,y,\theta+\epsilon\cdot v)\right)\right\} $.
    \State Find the inflection point as $\epsilon^\star\gets \text{Root} \left(\frac{d^2}{d\epsilon^2}f(\epsilon)=0\right) $ 

\State \Return $\theta_0+\epsilon^\star\cdot v$
\end{algorithmic}
\end{algorithm}

\subsubsection{Upper-bounding Pairwise Gradient Distances }
Next, we show that bounded loss difference at two perturbed checkpoints along some direction $v$ upper-bounds their gradient difference projected onto $v$. Crucially, this upper-bound is \textit{tightest} when the first checkpoint is the pretrained model and the other is the inflection point along direction $v$.

\begin{theoremE}[][end, no restate, text link=]
\label{thm:loss=grad}
Assume a smooth fine-tuning loss
with bounded curvature $0\le c\le c_H$. Starting from initialization $\theta_0$, for any  checkpoint $\theta \in \Theta$ during training define $\epsilon_{\theta} = \|\theta - \theta_0\|_2.$
Assume that fine-tuning gradients are predominantly captured by an $r$-dimensional subspace $\mathcal{V} \subset \mathbb{R}^d$ with orthonormal basis $\{v_1,\dots,v_r\}$, such that for every training example $z$,  
$
\|\text{Proj}_{\mathcal{V}^\perp}\nabla \mathcal{L}_z(\theta)\| \le \gamma.
$
For each basis direction $v_i$, denote $c_{H,i} \triangleq c_{H,v_i}$, and let
\[
\theta_{i,1} = \theta_0 + \epsilon_{i,1} v_i,\qquad
\theta_{i,2} = \theta_0 + \epsilon_{i,2} v_i, \qquad \text{where} ~~ 0 \le \epsilon_{i,1} < \epsilon_{i,2}.
\]

If any pair of training examples $z_1$ and $z_2$ have bounded loss difference at all checkpoints $\theta_{i,1}, \theta_{i,2}$:
\[
|\mathcal{L}_{z_1}(\theta_{i,1}) - \mathcal{L}_{z_2}(\theta_{i,1})| \le \delta,\quad
|\mathcal{L}_{z_1}(\theta_{i,2}) - \mathcal{L}_{z_2}(\theta_{i,2})| \le \delta,\quad
\forall i \in \{1,\dots,r\},
\]
then for any $\theta \in \Theta$ during training, the (full-rank) gradient difference is bounded by
\begin{equation}
\|\nabla \mathcal{L}_{z_1}(\theta) - \nabla \mathcal{L}_{z_2}(\theta)\|
\le
\sqrt{\sum_{i=1}^r \left( \frac{2\delta}{\epsilon_{i,2} - \epsilon_{i,1}} + \frac{1}{2}(\epsilon_{i,1} + \epsilon_{i,2})c_{H,i} + 2\epsilon_{\theta}\sqrt{c_H c_{H,i}} \right)^2} + 2\gamma.
\nonumber
\end{equation}
\end{theoremE}
\begin{proofE}
Let $\theta \in \Theta$ be an arbitrary parameter vector. Let $f(\theta) \triangleq \mathcal{L}_{z_1}(\theta) - \mathcal{L}_{z_2}(\theta)$ be the difference in loss of the two data points.
Let $g(\theta) \triangleq \nabla f(\theta) = \nabla \mathcal{L}_{z_1}(\theta) - \nabla \mathcal{L}_{z_2}(\theta)$ and $H(\theta) \triangleq \nabla^2 f(\theta)= \nabla^2 \mathcal{L}_{z_1}(\theta) - \nabla^2 \mathcal{L}_{z_2}(\theta)$ be the difference of gradient and the Fessian of $f$, respectively.
Define
\[
v_{\theta} \triangleq \frac{\theta - \theta_0}{\|\theta - \theta_0\|_2}.
\]

Fix any basis direction $v_i$. Let $\theta_{i,1} = \theta_0 + \epsilon_{i,1} v_i$ and $\theta_{i,2} = \theta_0 + \epsilon_{i,2} v_i$.
Let $u_{i,1} = \theta_{i,1} - \theta$ be the displacement vector from $\theta$ to the first perturbed checkpoint. Then $\theta_{i,2} - \theta = u_{i,1} + (\epsilon_{i,2} - \epsilon_{i,1}) v_i$. Expanding $f(\theta_{i,1})$ and $f(\theta_{i,2})$ around $\theta$ via Taylor series:
\begin{align*}
f(\theta_{i,1}) &= f(\theta) + \langle g(\theta), u_{i,1} \rangle + \frac{1}{2} u_{i,1}^\top H(\theta) u_{i,1} + R_{i,1} \\
f(\theta_{i,2}) &= f(\theta) + \langle g(\theta), u_{i,1} + (\epsilon_{i,2} - \epsilon_{i,1}) v_i \rangle + \frac{1}{2} (u_{i,1} + (\epsilon_{i,2} - \epsilon_{i,1}) v_i)^\top H(\theta) (u_{i,1} + (\epsilon_{i,2} - \epsilon_{i,1}) v_i) + R_{i,2},
\end{align*}
where we assume the Lagrange remainder terms are negligible.
Thus, subtracting the first equation from the second eliminates $f(\theta)$, $\langle g(\theta), u_{i,1} \rangle$, and the $\frac{1}{2} u_{i,1}^\top H(\theta) u_{i,1}$ term:
\begin{align*}
f(\theta_{i,2}) - f(\theta_{i,1}) = (\epsilon_{i,2} - \epsilon_{i,1})\langle g(\theta), v_i \rangle + (\epsilon_{i,2} - \epsilon_{i,1})u_{i,1}^\top H(\theta) v_i + \frac{1}{2}(\epsilon_{i,2} - \epsilon_{i,1})^2 v_i^\top H(\theta) v_i
\end{align*}
Rearranging gives:
\begin{equation*}
\langle g(\theta), v_i \rangle = \frac{f(\theta_{i,2}) - f(\theta_{i,1})}{\epsilon_{i,2} - \epsilon_{i,1}} - u_{i,1}^\top H(\theta) v_i - \frac{1}{2}(\epsilon_{i,2} - \epsilon_{i,1}) v_i^\top H(\theta) v_i
\end{equation*}
Applying triangle inequality:
\begin{equation*}
|\langle g(\theta), v_i \rangle| \le \frac{|f(\theta_{i,2}) - f(\theta_{i,1})|}{\epsilon_{i,2} - \epsilon_{i,1}} + |u_{i,1}^\top H(\theta) v_i| + \frac{1}{2}(\epsilon_{i,2} - \epsilon_{i,1}) |v_i^\top H(\theta) v_i|
\end{equation*}

Given $|f(\theta_{i,1})| \le \delta$ and $|f(\theta_{i,2})| \le \delta$, we have $|f(\theta_{i,2}) - f(\theta_{i,1})| \le 2\delta$.
\[
u_{i,1} = \theta_{i,1} - \theta = (\theta_0 + \epsilon_{i,1}v_i) - (\theta_0 + \epsilon_{\theta}v_{\theta}) = \epsilon_{i,1}v_i - \epsilon_{\theta}v_{\theta}.
\]
Therefore,
\begin{equation*}
\left|u_{i,1}^\top H(\theta) v_i\right|
\le
\epsilon_{i,1}\left|v_i^\top H(\theta) v_i\right|
\,+\,
\epsilon_{\theta}\left|v_{\theta}^\top H(\theta) v_i\right|.
\end{equation*}
For the first term, since $0 \le v_i^\top \nabla^2 \mathcal{L}_{z_j}(\theta) v_i \le c_{H,i}$ for $j\in\{1,2\}$,
\begin{equation*}
\left|v_i^\top H(\theta) v_i\right|
= \left|v_i^\top \nabla^2 \mathcal{L}_{z_1}(\theta) v_i - v_i^\top \nabla^2 \mathcal{L}_{z_2}(\theta) v_i\right|
\le c_{H,i}.
\end{equation*}
For the mixed term, by triangle inequality and generalized Cauchy-Schwarz for PSD matrices,
\begin{equation*}
\left|v_{\theta}^\top H(\theta) v_i\right|
\le
\sum_{j=1}^2 \left|v_{\theta}^\top \nabla^2 \mathcal{L}_{z_j}(\theta) v_i\right|
\le
\sum_{j=1}^2 \sqrt{v_{\theta}^\top \nabla^2 \mathcal{L}_{z_j}(\theta) v_{\theta}}\sqrt{v_i^\top \nabla^2 \mathcal{L}_{z_j}(\theta) v_i}
\le
2\sqrt{c_H c_{H,i}}.
\end{equation*}
Thus,
\begin{equation*}
\left|u_{i,1}^\top H(\theta) v_i\right|
\le
\epsilon_{i,1}c_{H,i} + 2\epsilon_{\theta}\sqrt{c_H c_{H,i}}.
\end{equation*}
Also, $\left|v_i^\top H(\theta) v_i\right|\le c_{H,i}$, so
\begin{equation*}
|\langle g(\theta), v_i \rangle|
\le
\frac{2\delta}{\epsilon_{i,2} - \epsilon_{i,1}}
\,+\,
\epsilon_{i,1}c_{H,i}
\,+\,
2\epsilon_{\theta}\sqrt{c_H c_{H,i}}
\,+\,
\frac{1}{2}(\epsilon_{i,2} - \epsilon_{i,1})c_{H,i}
\triangleq B_i.
\end{equation*}
Equivalently,
\[
B_i
=
\frac{2\delta}{\epsilon_{i,2} - \epsilon_{i,1}}
\,+\,
\frac{1}{2}(\epsilon_{i,1} + \epsilon_{i,2})c_{H,i}
\,+\,
2\epsilon_{\theta}\sqrt{c_H c_{H,i}}.
\]

Since $\{v_1,\dots,v_r\}$ is an orthonormal basis of $\mathcal{V}$:
\begin{equation*}  
\|\text{Proj}_{\mathcal{V}}(g(\theta))\| = \sqrt{\sum_{i=1}^r \langle g(\theta), v_i \rangle^2} \le \sqrt{\sum_{i=1}^r B_i^2}.
\end{equation*}
By assumption on orthogonal components:
\begin{equation*}
\|\text{Proj}_{\mathcal{V}^\perp}(g(\theta))\|
\le
\|\text{Proj}_{\mathcal{V}^\perp}\nabla \mathcal{L}_{z_1}(\theta)\| +
\|\text{Proj}_{\mathcal{V}^\perp}\nabla \mathcal{L}_{z_2}(\theta)\|
\le 2\gamma.
\end{equation*}
Finally,
\begin{align*}
\|g(\theta)\|
&\le \|\text{Proj}_{\mathcal{V}}(g(\theta))\| + \|\text{Proj}_{\mathcal{V}^\perp}(g(\theta))\| \\
&\le \sqrt{\sum_{i=1}^r \left( \frac{2\delta}{\epsilon_{i,2} - \epsilon_{i,1}} + \frac{1}{2}(\epsilon_{i,1} + \epsilon_{i,2})c_{H,i} + 2\epsilon_{\theta}\sqrt{c_H c_{H,i}} \right)^2} + 2\gamma.
\end{align*}
\end{proofE}

\begin{corollaryE}[][end,text link=]
\label{coro:minimize}
The RHS upper-bound of the gradient difference in \Cref{thm:loss=grad} is tightest at: 
\[
\epsilon_{i,1}^\star = 0,\qquad
\epsilon_{i,2}^\star = 2\sqrt{\frac{\delta}{c_{H,i}}}.
\]
At this minimizer, the full-parameter gradient difference is bounded by:
\[
\|\nabla \mathcal{L}_{z_1}(\theta) - \nabla \mathcal{L}_{z_2}(\theta)\|
\le
2\left(\sqrt{\delta} + \epsilon_{\theta}\sqrt{c_H}\right)\sqrt{\sum_{i=1}^r c_{H,i}} + 2\gamma.
\]
\end{corollaryE}
\begin{proofE}
From \Cref{thm:loss=grad}, the directional term can be written as
\begin{equation*}
B_i(\epsilon_{i,1},\epsilon_{i,2})
=
\frac{2\delta}{\epsilon_{i,2}-\epsilon_{i,1}}
\,+\,
\frac{1}{2}(\epsilon_{i,1}+\epsilon_{i,2})c_{H,i}
\,+\,
2\epsilon_{\theta}\sqrt{c_H c_{H,i}}.
\end{equation*}
First, for fixed $\epsilon_{i,2}$,
\begin{equation*}
\frac{\partial B_i}{\partial \epsilon_{i,1}}
=
\frac{2\delta}{(\epsilon_{i,2}-\epsilon_{i,1})^2}
\,+\,
\frac{1}{2}c_{H,i}
>0,
\end{equation*}
so $B_i$ is strictly increasing in $\epsilon_{i,1}$ and is minimized at $\epsilon_{i,1}^\star=0$.
Substituting $\epsilon_{i,1}=0$ gives
\begin{equation*}
B_i(0,\epsilon_{i,2})
=
\frac{2\delta}{\epsilon_{i,2}}
\,+\,
\frac{1}{2}\epsilon_{i,2}c_{H,i}
\,+\,
2\epsilon_{\theta}\sqrt{c_H c_{H,i}}.
\end{equation*}
Now differentiate with respect to $\epsilon_{i,2}$:
\begin{equation*}
\frac{\partial B_i(0,\epsilon_{i,2})}{\partial \epsilon_{i,2}}
=
-\frac{2\delta}{\epsilon_{i,2}^2}
\,+\,
\frac{1}{2}c_{H,i}
=
0,
\end{equation*}
which yields
\begin{equation*}
\epsilon_{i,2}^\star
=
2\sqrt{\frac{\delta}{c_{H,i}}}.
\end{equation*}
At this minimizer,
\begin{equation*}
B_i^\star
=
\frac{2\delta}{2\sqrt{\delta/c_{H,i}}}
\,+\,
\frac{1}{2}\!\left(2\sqrt{\frac{\delta}{c_{H,i}}}\right)\!c_{H,i}
\,+\,
2\epsilon_{\theta}\sqrt{c_H c_{H,i}}
=
2\sqrt{\delta c_{H,i}}
\,+\,
2\epsilon_{\theta}\sqrt{c_H c_{H,i}}.
\end{equation*}
Substitute into \Cref{thm:loss=grad}:
\begin{align*}
\|\nabla \mathcal{L}_{z_1}(\theta) - \nabla \mathcal{L}_{z_2}(\theta)\| &\le \sqrt{\sum_{i=1}^r B_i^2} + 2\gamma \\
&= \sqrt{\sum_{i=1}^r \left(2\sqrt{c_{H,i}}\left(\sqrt{\delta}+\epsilon_{\theta}\sqrt{c_H}\right)\right)^2} + 2\gamma \\
&= 2\left(\sqrt{\delta}+\epsilon_{\theta}\sqrt{c_H}\right)\sqrt{\sum_{i=1}^r c_{H,i}} + 2\gamma.
\end{align*}
\end{proofE}

\textbf{\textit{Remark.}} \Cref{coro:minimize} suggests that to get the tightest upper-bound on pairwise gradient differences, LoRA directions with smaller curvature should receive larger perturbations (i.e., up to their inflection point). 

\begin{figure}[t]
    \centering
    \includesvg[width=\linewidth]{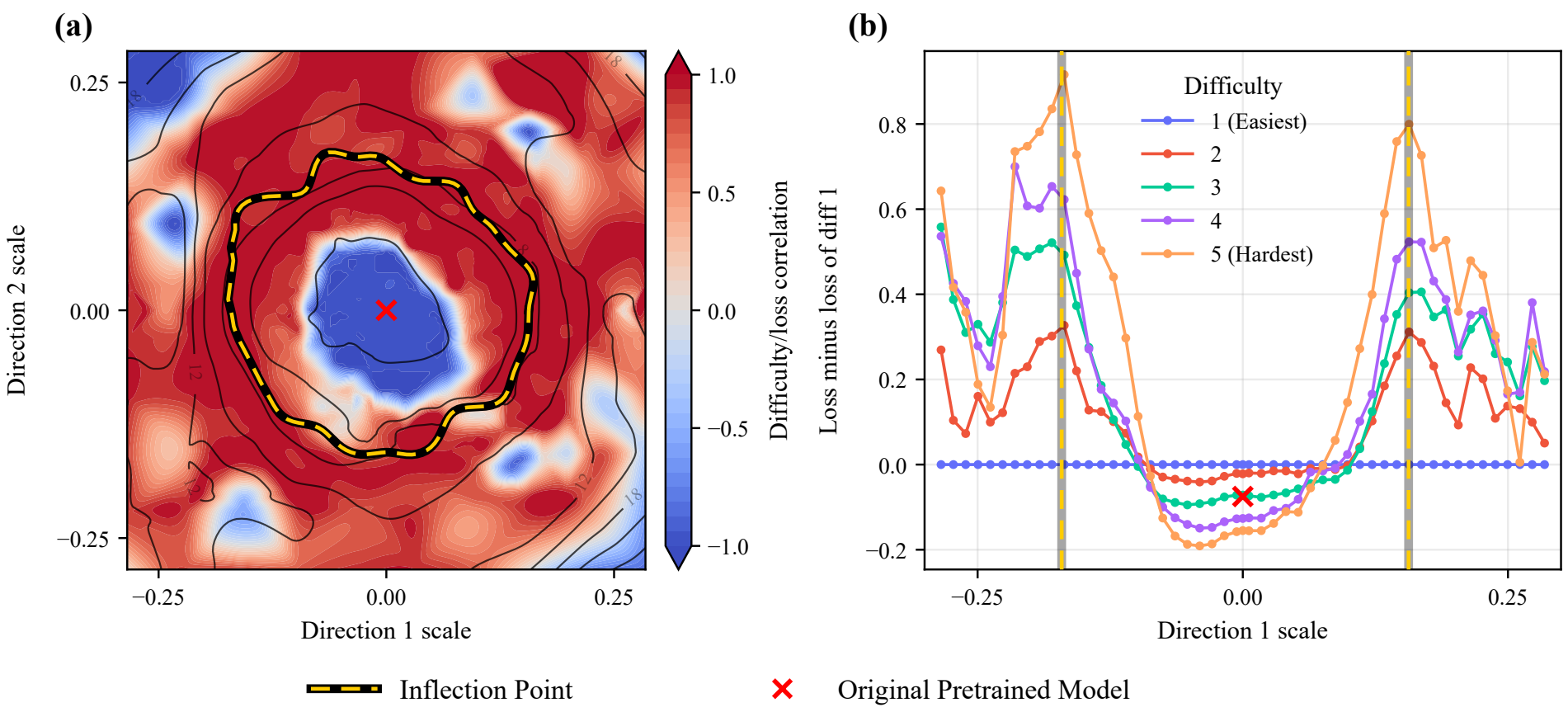}
    \vspace{-7mm}
    \caption{
(left) correlation between the initial 100 token and the problem difficulty in the 2D contour plot of the loss. While the loss evaluated at the pretrained model does not have a good correlation with difficulty (blue region near the origin); the loss of perturbed parameters correlates well with difficulty (red region outside the blue circle). 
Crucially, the \textit{best separation} of difficulty levels occurs at the \textit{inflection} point (yellow band). (right) examples that are labeled more difficult by LLM have a higher curvature for their initial 100 token loss around the pretrained model parameters. }
    \label{fig:loss_difficulty}
\end{figure}

\subsubsection{Diversity Filtering} 
To predict which reasoning traces will be diverse if fully generated, we first find the fixed low-dimensional subspace $\mathcal{V}$ by finetuning the model on a small subset of the data using LoRA with a small rank $r$. 
The use of LoRA here is merely for finding the dominant subspace, and we are not restricted to LoRA during the actual SFT. Performing QR decomposition on the LoRA adaptors gives us the orthogonal basis vectors $\{v_1, \dots, v_r\}$ of $\mathcal{V}$. 
Then, following \Cref{coro:minimize}, we find $\epsilon_{i,1},\epsilon_{i,2}$
in every direction $v_i$ using Algorithm \ref{alg:inflection-search}. 
This yields two checkpoints for each direction $v_i$. 
Following \citep{ICML}, we only generate and evaluate the loss of the initial 1k tokens of every reasoning trace on the $2r$ checkpoints to get the \textit{loss signature} for every example $j$:
\begin{equation}\label{eq:signature}
\hspace{-2mm} \text{\textit{\textbf{\textbf{Loss signature:}}}} ~   L_j = [\mathcal{L}_j^{\text{1k}}(\theta_0+\epsilon_{1,1}v_1), \mathcal{L}_j^{\text{1k}}(\theta_0+\epsilon_{1,2}v_1),\cdots, \mathcal{L}_j^{\text{1k}}(\theta_0+\epsilon_{r,1}v_r), \mathcal{L}_j^{\text{1k}}(\theta_0+\epsilon_{r,2}v_r)]
\end{equation}
Theoretically, $\epsilon^*_{i,1}=0$ for all $i$. To avoid all $\theta_0+\epsilon_{i,1}v_i$ collapsing to $\theta_0$ and have more informative loss signatures, we perturb $\epsilon_{i,1}$ until it satisfies:
\begin{equation}
    \E_j\left[\mathcal{L}_j^{\text{1k}}(\theta_0+\epsilon_{i,1}v_i)\right] = ({\E_j\left[\mathcal{L}_j^{\text{1k}}(\theta_0)\right]+\E_j\left[\mathcal{L}_j^{\text{1k}}(\theta_0+\epsilon_{i,2}v_i)\right]})/{2},
    \label{eq:eps1}
\end{equation}
and use the obtained $\epsilon_{i,1}$ for every $v_i$ to calculate the loss signatures. Detailed implementation is in \Cref{alg:lora-train-perturb}.

\textbf{Clustering loss signatures.} We standardize each samples to zero mean and unit variance, and use FAISS \citep{douze2024faiss} to cluster the data points into $k$ clusters. We then select an equal number of data points from each cluster to ensure that the data points are diverse. Empirical gains of diversity filtering is presented in \Cref{app:emp-div}.

\begin{algorithm}[t]
    \caption{Complete data curation pipeline}
    \label{alg:pipeline-short-short}
    \begin{algorithmic}[1]
    \Require  Mixture of problems from different sources
    \Ensure High-quality SFT data
    \State Select the teacher model according to \Cref{sec:teacher-selection} 
    \State Generate the initial 100 reasoning tokens for all the examples
    \State Calculate problem difficulties in each source based on the first 100 tokens (Eq. \ref{eq:difficulty}) 
    \State Cluster examples in each source based on their difficulty scores into two clusters and drop the easy cluster with lower loss
    \State Generate the initial 1k reasoning tokens for the remaining examples
    \State Calculate loss signature (Eq. \ref{eq:signature}) in every source based on the initial 1k tokens and cluster them to find similar reasoning traces
    \State Sample a balanced subset of $N_s$ (Eq.  \ref{eq:budget}) examples from clusters in every source
    \State Generate full reasoning traces for the sampled examples
    \end{algorithmic}
\end{algorithm}
\subsection{Predicting Reliable Difficulty Scores}
\label{subsec:difficulty}
Next, we discuss predicating reliable difficulty scores for all examples. When generating reasoning traces, the teacher model starts its reasoning by summarizing and rephrasing the problem. Thus, the loss of the initial 100 tokens 
along the training trajectory provides high-quality signal for problem difficulty \citep{ICML}. Here, we show how reliable difficulty scores can be calculated without having to train the model on full reasoning traces to obtain the training trajectory. 

\textbf{Difficult problems induce higher loss curvature.} Our key observation is that examples with different difficulty levels have different loss curvature for their initial 100 tokens. As shown in \Cref{fig:loss_difficulty} (right), examples that are labeled more difficult by LLM have a higher curvature for their initial 100 token loss around the pretrained model parameters. 
\Cref{fig:loss_difficulty} (left) puts the correlation between this loss on individual problem and the difficulty of that problem in the 2D contour plot of the loss. We see that while the loss evaluated at the pretrained model does not have a good correlation with the difficulty of the problem (blue region near the origin); once perturbed, the loss correlates well with the difficulty of the problem (red region outside the blue circle). 
Crucially, the \textit{best separation} of difficulty levels occurs at the \textit{inflection} point along a random direction $\hat u$ found by Algorithm \ref{alg:inflection-search}. 
\begin{equation}\label{eq:difficulty}
    \hspace{-1cm}\text{\textit{\textbf{\textbf{Difficulty score:}}}} \quad\quad d_j = \mathcal{L}_j^{\text{100}}(\theta_0 + \epsilon_{i,2}\hat u),
\end{equation}
Unlike diversity filtering, we do not require the LoRA directions and any random direction $\hat u$ can be used for calculating difficulty scores.

\subsubsection{Difficulty Filtering}
SFT data is often curated using mixture of problems from different sources. 
For example, NuminaMath \citep{numinamath}, which is a mathematical reasoning dataset contains international Olympiad problems, high school math, math exams, online forums \& repositories, and synthetic data. 

\textbf{Filtering easy problems in every source.} For problems in every source, we calculate their difficulty scores using Eq. \ref{eq:difficulty}. Then, we use FAISS \citep{douze2024faiss} to cluster difficulty scores in each source into two clusters and discard the easy cluster with lower average loss.

\textbf{Sampling more from more difficult sources.} Next, we calculate the average difficulty of problems $d_j$ remained in every source $V_s$. 
For a total budget of $N$ data examples to curate, we sample more examples from sources with higher average difficulty, as follows:
\begin{equation}\label{eq:budget}
    N_s=N \cdot \text{softmax}_s(\mathbb{E}_{j \in V_s}[d_j]).
\end{equation}
The complete pseudocode of the method is illustrated in Alg. \ref{alg:pipeline-short-short} with further detail in Alg. \ref{alg:pipeline}.

\section{Teacher Selection}
\label{sec:teacher-selection}
Finally, we describe how to select the best teacher model for a student, without having to train the model or generate full reasoning traces.

\begin{figure}
    \centering
    \includesvg[width=0.9\linewidth]{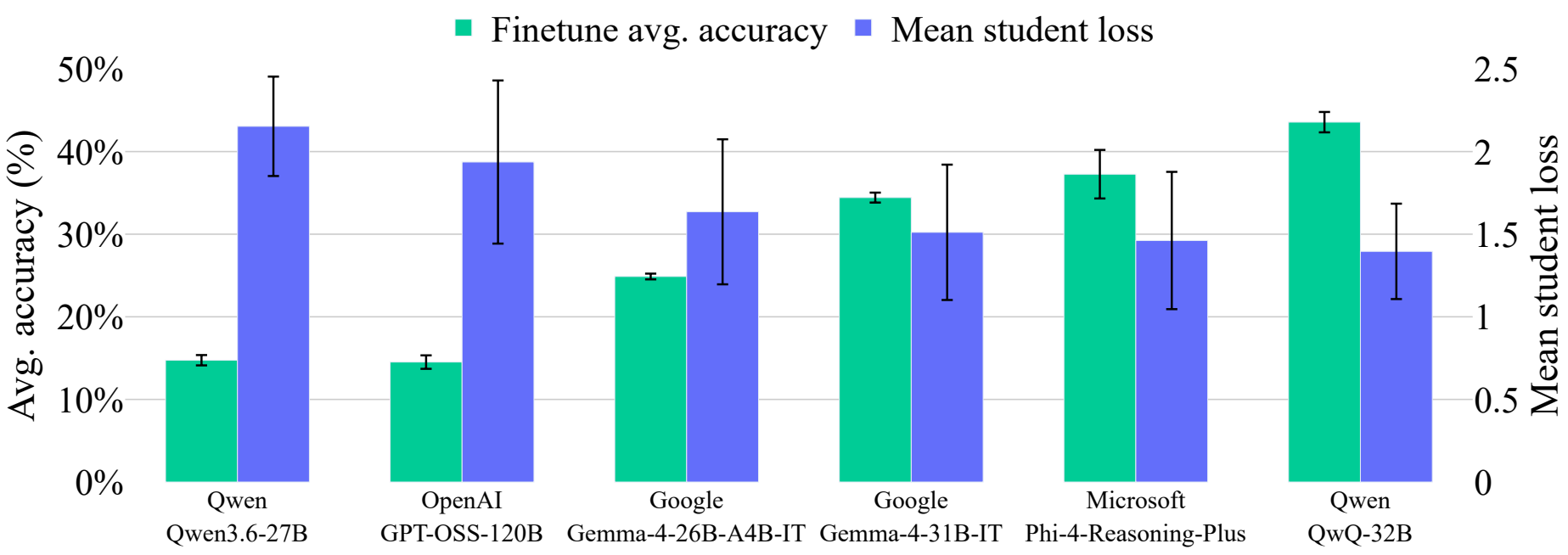}
    \vspace{-6mm}
    \caption{We evaluate the loss of the student model and the accuracy of the finetuned on the dataset generated from the teacher model on the NuminaMath dataset. We also train the student using the same randomly selected 1000 questions with responses generated from the teacher model. We find that the higher the rephrasing loss is, the worse the final finetuned student performs.}
    \label{fig:teacher-selection}
\end{figure}

\textbf{Student learns better from teachers with lower rephrasing loss.} We aim to select the teacher whose language is most familiar to the student, allowing the student to focus on solving the problem rather than parsing unfamiliar phrasing. To this end, we randomly sample a set of questions and prompt each candidate teacher model to rephrase the same problems. We cap each response at 100 tokens, as in difficulty filtering, and evaluate the rephrasings using the \textit{pretrained} student parameters rather than the perturbed ones. Specifically, let $y_{ij}^{rephr}$ denote the rephrasing generated by teacher model $i$ for problem $j$. We define its rephrasing loss under the pretrained student model $\theta_0$ as:
\begin{equation}\label{eq:teacher-rephrasing}
    \hspace{-1cm}\text{\textit{\textbf{\textbf{Rephrasing Loss:}}}} \quad\quad R_i=\E_j\left[\mathcal{L}^{\text{100}}(x_j,y_{ij}^{rephr},\theta_0)\right],
\end{equation}
As shown in \Cref{fig:teacher-selection}, the teachers with the lowest average loss can best train the student model. Additional details and discussions are provides in \Cref{app:teacher}.

We emphasize the distinction between the loss on individual problems and the average loss across problems. For a fixed set of problems, varying the teacher allows us to identify the teacher whose rephrasings are most compatible with the student’s existing language patterns. A teacher with lower loss produces phrasing that the pretrained student already understands well, while preserving the underlying problem to be solved. This allows the student to devote its capacity to learning the solution rather than adapting to a new linguistic pattern.
Conversely, a high loss indicates that the student struggles to interpret the teacher’s rephrasing. Training on such teachers can therefore push the student away from its pretrained optimum, potentially degrading existing capabilities or even inducing catastrophic forgetting. This interpretation is consistent with observations in \cite{panigrahi2025good,teacher-selection}.

\section{Experiments}
In this section, we evaluate the performance of \method\ on the following tasks:
\begin{enumerate}[topsep=-1pt,itemsep=0ex,partopsep=1ex,parsep=1ex,leftmargin=0.5cm]
    \item \textbf{Data selection for medical reasoning}: 
    First, we apply \method\  to \textit{select} 1k data points from the pre-generated M23k dataset \citep{M23k} and evaluate the performance of the student model finetuned on the selected data points. This serves as a demonstration for \method's applicability in the setting of data selection as well and to provide a controlled comparison against the data selection baselines. 
    \item \textbf{Data curation for medical reasoning}: Then, we use the original, unfiltered set of 37,698 decontaminated \textit{questions} given in \citet{M23k}---that is used to curate the M23k datase---to generate a new reasoning SFT data. We applied \method\ to select the teacher model (among the smaller models we could fit in our GPU memory) and QwQ-32B is selected. We generated a dataset containing 1000 reasoning examples using our method.
    \item \textbf{Data curation for math reasoning}: We generate reasoning traces from a subset of the NuminaMath dataset \citep{numinamath} containing 192k questions using the best teacher model selected by \Cref{sec:teacher-selection} and evaluate the performance of the student model finetuned on the generated dataset. We compare with OpenThinker-7B \citep{openthoughts} and other data selection methods that select data from OpenThoughts-Math, a reasoning dataset generated from the same subset of NuminaMath.
    \item \textbf{Data curation for physics reasoning}: We sanitized the problems and ground-truth solutions from several physics olympics from 1967 to 2018 into 3084 pairs of problems and solutions. We combine it with physics problems from PhysicsEval \citep{physicsEval} to form a mixed dataset of 40k problems and curate a physics reasoning dataset.
\end{enumerate}
Additional details on datasets, benchmarks, baselines and hyperparameters are in \Cref{app:add-expr}.\looseness=-1

\subsection{Results}
\begin{figure}
    \centering
    \includesvg[width=1\textwidth]{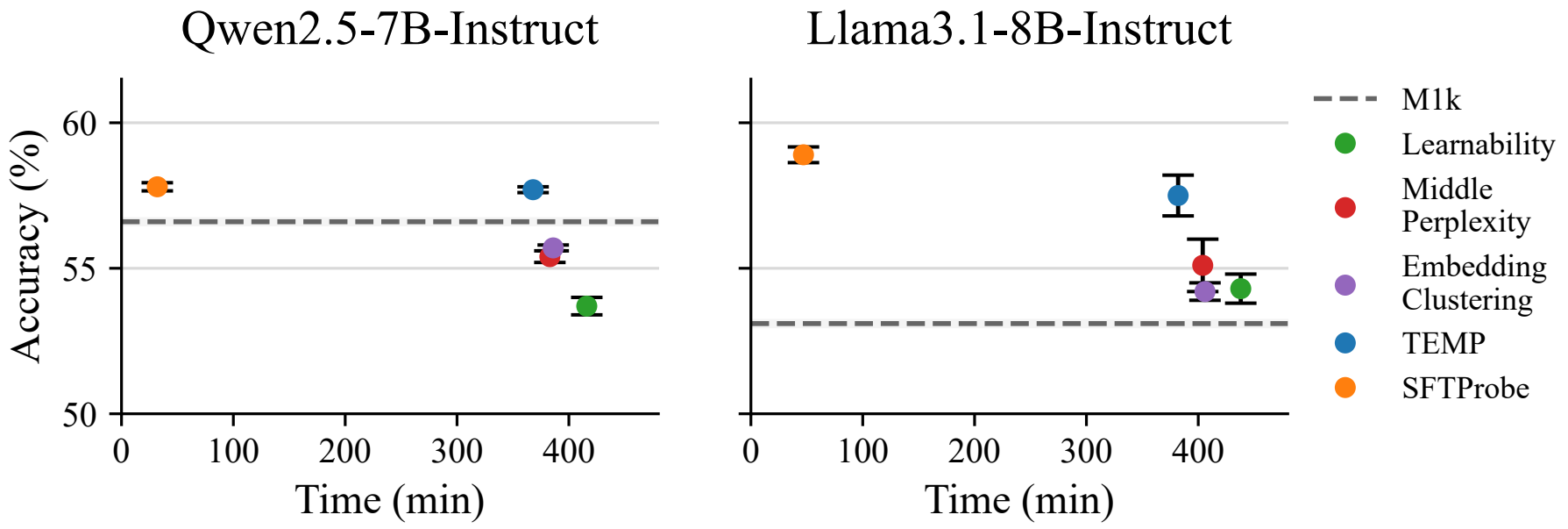}
    \vspace{-10mm}
    \caption{Results on data selection for medical reasoning with different models as the base model. Wall time on 4$\times$H100 is reported. \method\ performs comparable to best selection baseline on Qwen2.5-7B and outperforms selection baselines on Llama3.1-8B, while being substantially faster.}
    \label{fig:M1k-results}
    \vspace{-3mm}
\end{figure}

\begin{figure}
    \centering
    \includesvg[width=0.6\textwidth]{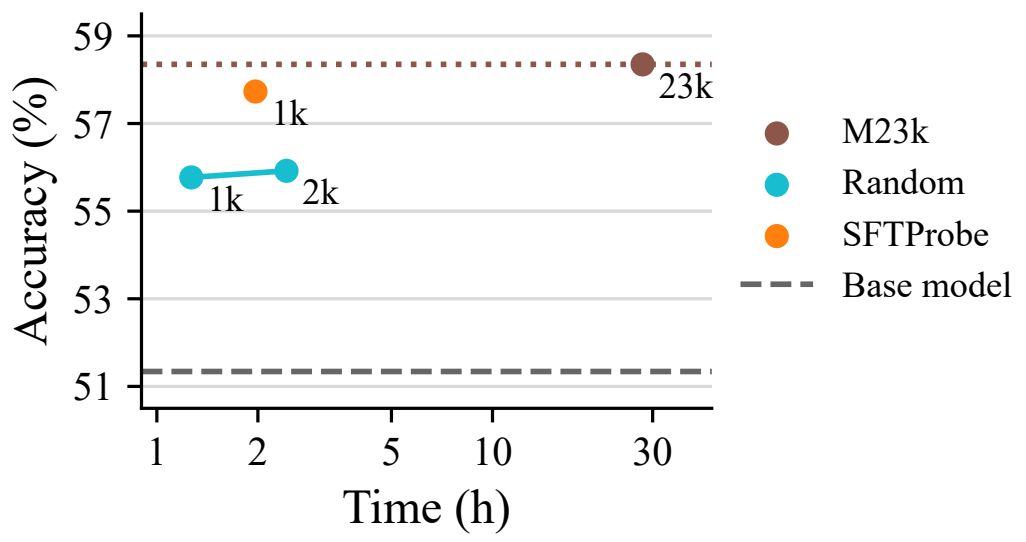}
    \vspace{-2mm}
    \caption{Results on data curation for medical reasoning. Wall time on 4$\times$H100 is reported. Compared to the full 23k data finetune, our curated 1k data closes 76\% of the gap between random sampling and full dataset training, while only requiring 7\% of the compute cost.}
    \label{fig:medical-curation}
\end{figure}

\begin{figure}
    \centering
    \includesvg[width=0.6\textwidth]{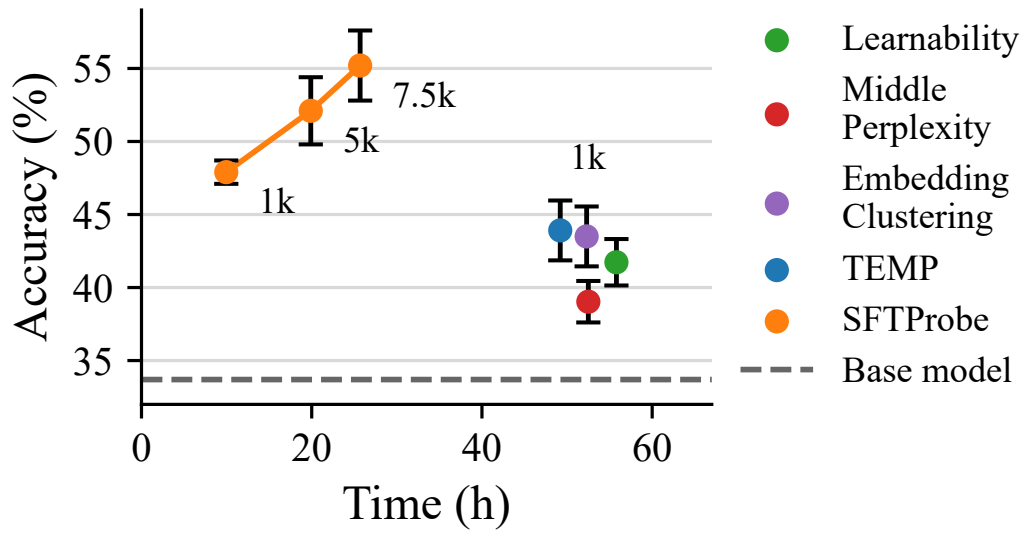}
    \vspace{-2mm}
    \caption{Results on data curation for math reasoning. Data selection methods select 1k data from Openthoughts-Math dataset containing pre-generated teacher examples. \method\  generates data with QwQ-32B as the teacher, and even when taken into consideration the generation time, takes less time compared to the data selection baselines. Wall time on 4$\times$H100 is reported.}
    \label{fig:math-results}
    \vspace{-3mm}
\end{figure}

Our data curation pipeline provides substantial increment in performance across multiple datasets. When selecting from generated reasoning traces, given the same data budget, we're see a 1.1\% increase in performance in M23k dataset in \Cref{fig:M1k-results} when using Llama3.1-8B-Instruct model compared to the best baseline. When the reasoning traces are curated, training on 1k data curated with \method\  matches the full curated M23k dataset within 0.6\% as shown in \Cref{fig:medical-curation}. On math reasoning, we see a 3.4\% increase in performance in math reasoning in \Cref{fig:math-results} compared to the best baseline. Evaluation on the physics dataset in \Cref{tab:physics_results} shows a performance close to PRIME-RL/P1-30B-A3B, a model capable of winning the silver model for IPhO 2025. With a higher budget, we curated a 5k math reasoning dataset. Using the same student model, we match OpenThinker-7B as shown in \Cref{fig:intro-numina}, which used 89k math reasoning data to train.

Apart from effective, \method\  is also highly efficient. We were able to obtain the same math performance using 53x less computation compared to OpenThinker-7B as shown in \Cref{fig:intro-numina}. \method\ even requires less time to run compared to selection methods where the long reasoning traces have aleady been generated, as shown in \Cref{fig:math-results}.


\begin{table}[t]
    \centering
    \caption{Experiment results on data curation for physics reasoning with Qwen2.5-7B-Instruct as the student model. We outperform PRIME-RL/P1-30B-A3B, a model capable of winning the silver model for IPhO 2025, although we start with a much weaker base model.}
    \label{tab:physics_results}
    
    \begin{tabular}{rc}
        \toprule
        Model & Average Score \\
        \midrule
        Qwen2.5-7B-Instruct & 3.58 \\
        Google/Gemma-4-26B-A4B-it & 9.65 \\
        PRIME-RL/P1-30B-A3B \citep{p1-2025} & 9.05 \\
        \midrule
        \method\  (1k data generated) & 9.14 $\pm$ 0.29 \\
        \method\  (2k data generated) & 9.32 $\pm$ 0.11 \\
        \method\  (5k data generated) & 9.76 $\pm$ 0.74 \\
        \bottomrule
        \end{tabular}
\end{table}

\subsection{Ablation Studies}
In this section, we conduct ablation studies to demonstrate the insensitivity of the hyperparameters and the effects of individual components in \method.
\vspace{-2mm}
\paragraph{\textit{LoRA Rank.}} The most significant hyperparameter in \method\  is the rank of the brief LoRA training, which controls the dimension of the subspace and the length of the loss signatures. In theory, LoRA automatically finds the most important dimensions so that increasing the rank has diminishing returns. This is also supported by the \Cref{fig:loss-3d}(b) and (c), which demonstrates that the model is significantly more dependent on the first few dimensions. Results in \Cref{tab:ablations} demonstrate that reducing rank from 4 to 2 does not significantly affect performance.
\vspace{-2mm}
\paragraph{\textit{LoRA Training Steps.}} Unlike \citet{ICML,S2L} that requires a full training run to get the training trajectory, \method\  relaxes this assumption and works well even with just a few iterations of LoRA training. \Cref{tab:ablations} show that using only 100 - 400 iterations is sufficient to identify the dominant training directions.
\vspace{-2mm}
\paragraph{\textit{Individual steps.}} We ablate the effectiveness of different components of \method\  in \Cref{tab:M23k_results-qwen}. For diversity filtering, replacing the inflection points with checkpoints along the training trajectory leads to a $0.7\%$ decrease in performance. Furthermore, using checkpoints along the brief (approximate) training trajectory used in our LoRA training (instead of checkpoints along the full training trajectory) leads to additional $0.8\%$ decrease in accuracy.
\begin{table}[t]
    \centering
    \caption{Ablation Study for the effect of inflection search and brief LoRA training.}
    \label{tab:M23k_results-qwen}
    \vspace{-2mm}
    \begin{tabular}{lccc}
    \toprule
    \multirow{2}{*}{Method} & ID & OOD & Avg \\
    & (3 sets) & (7 sets) & (10 sets) \\
   \midrule
   \citet{ICML}  & \textbf{69.3\%} & 52.9\% & \textbf{57.8\%} \\\midrule
   \method\    & 68.7\% & \textbf{53.1\%} & \textbf{57.8\%} \\
   ~ without Inflection search & 67.9\% & 52.5\%& 57.1\%\\ 
   ~ without LoRA & 68.1\% & 51.3\% & 56.3\% \\
    \bottomrule
    \end{tabular}
\end{table}
\begin{table}[H]
    \centering
    \caption{Ablation study on the key hyperparameters. We vary the number of training steps and LoRA rank and evaluate the performance of the student model finetuned on the generated dataset. A small number of training steps and LoRA rank is sufficient to capture the dominant directions of the fine-tuning loss landscape and yield satisfactory performance.}
    \label{tab:ablations}
    \begin{tabular}{rccc}
        \toprule
        \multirow{2}{*}{Hyperparameter} & ID & OOD & Avg \\
        & (3 sets) & (7 sets) & (10 sets) \\
       \midrule
       Lora Rank = 4, Train with 200 iters & 68.7\% & \textbf{53.1\%} & \textbf{57.8\%} \\
       Lora Rank = 2, Train with 200 iters & 68.5\% & 52.6\% & 57.3\% \\
    Lora Rank = 4, Train with 100 iters & 68.9\% & 51.9\% & 57.0\% \\
    Lora Rank = 4, Train with 400 iters & \textbf{69.0\%} & 52.0\% & 57.1\% \\
        \bottomrule
        \end{tabular}
\end{table}
\vspace{-4mm}
\section{Conclusion}\vspace{-2mm}
In this work, we introduced a novel paradigm that directly predicts the quality of reasoning data from local loss geometry. \method\ is based on the observation that pretrained LLMs operate within a low-rank, anisotropic loss basin. By briefly adapting LoRA directions and evaluating loss on only 5\% of reasoning tokens, we construct compact loss signatures that reliably estimate both diversity and difficulty. We further showed that clustering these signatures captures gradient similarity during SFT, while loss at curvature inflection points provides a robust proxy for difficulty. We also provide a simple criterion for teacher selection.
 
Empirically, \method\  reduces data curation time from weeks to minutes while maintaining or improving downstream performance. In mathematical reasoning, 5K selected examples match the heavily curated 89K-example OpenThoughts dataset, and we further introduce a high-quality physics reasoning dataset. Overall, our results suggest that reasoning data quality can be predicted rather than exhaustively searched, opening a scalable and principled direction for LLM post-training.

\section*{Acknowledgements}
This research was supported by Optum AI Science, NSF
CAREER Award 2146492, NSF-Simons AI Institute for
Cosmic Origins (CosmicAI), NSF AI Institute for Foundations of Machine Learning (IFML), and an Apple Research Award.

{\small\printbibliography}

\newpage
\appendix
\section{Proofs omitted in the main body}
\printProofs
\section{Mechanism of difficulty filtering}
\begin{figure}
    \centering
    \includesvg[width=\linewidth]{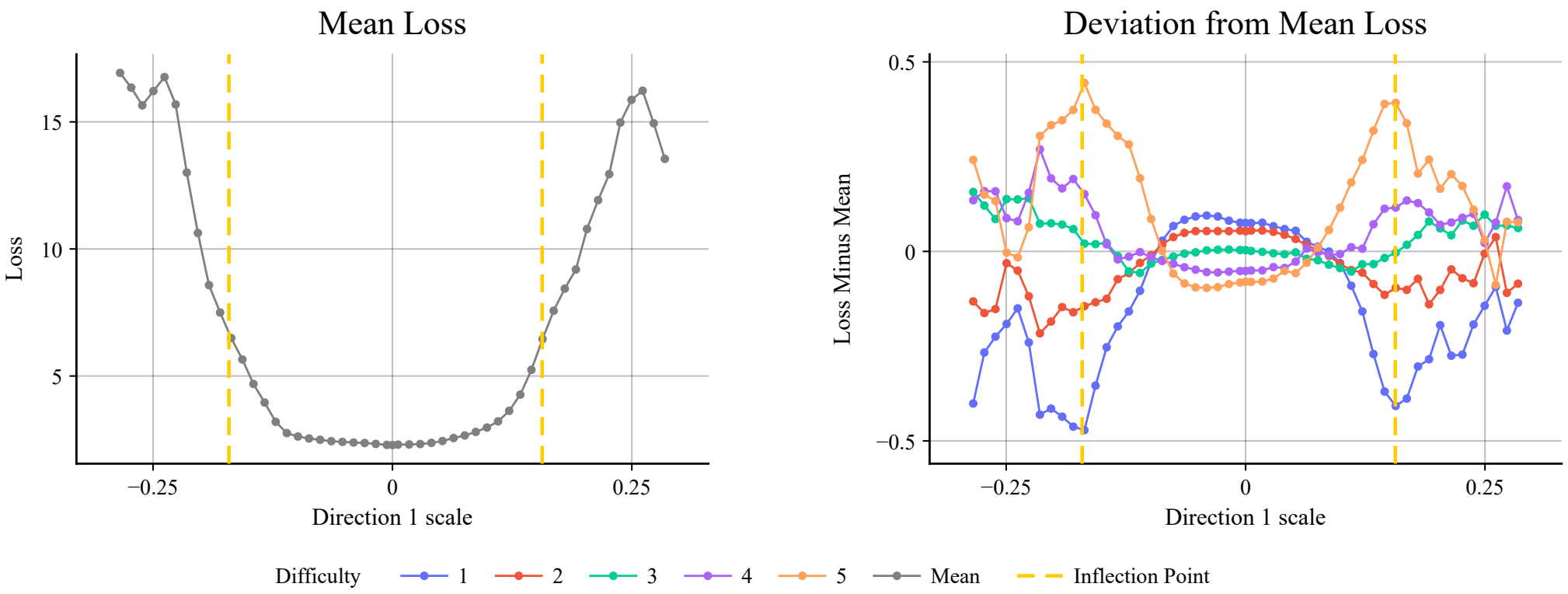}
    \caption{Left: The loss of Qwen2.5-3B-Instruct perturbed along a random direction on M23k dataset. Right: The average loss of problems in each difficulty level minus the average loss of all the problems. Within the pretrained model's proximity, the loss of difficult problems grows faster than the easy problems as the perturbation increases.}
    \label{fig:loss_by_difficulty}
\end{figure}

We provide further analysis on why the loss on perturbed model indicates difficulty of the problem.
We first provide additional context of \Cref{fig:loss_difficulty} (right) and \Cref{fig:loss_by_difficulty}. Both plots show the loss of the model Qwen2.5-3B-Instruct perturbed along random directions on the first 100 tokens of the M23k dataset to obtain $\mathcal{L}_{problem}^{100}(x,y,\theta_{perturbed})$. We labelled the difficulty of the problems in the dataset with Gemini-2.5-Flash into 5 levels, with level 1 being the easiest and level 5 being the hardest. The loss are grouped by difficulty level and averaged. We calculate the Pierson correlation coefficient between the loss and the difficulty level to obtain the correlation plot in \Cref{fig:loss_difficulty} (right). We calculate the average loss of all the problems to obtain the loss curve in \Cref{fig:loss_by_difficulty} (left). For \Cref{fig:loss_by_difficulty} (right), we calculated the average loss on each difficulty level and subtracted mean loss of all the problems.

\Cref{fig:loss_by_difficulty} provides insights into the mechanism of difficulty filtering. When we perturb the model, the loss doesn't increase to the same extent for easy and difficult problems. The subplot on the right shows that loss of easier problems grows slower than average while the loss of difficult problems grows faster as the perturbation increases while keeping the model close to the pretrained checkpoint (within the inflection points). Such difference magnifies the difference of loss between easy and difficult problems to up to 5x, making them more separable.
Therefore, within the proximity of the pretrained checkpoints (indicated by the inflection ring), the larger the perturbation is, the more separable the loss of easy and difficult problems are. We utilize this insight to use \Cref{alg:inflection-search} to find the appropriate perturbation scale for the difficulty filtering.

\begin{figure}
\centering
\vspace{-0.2cm}
\includesvg[width=1.15\linewidth]{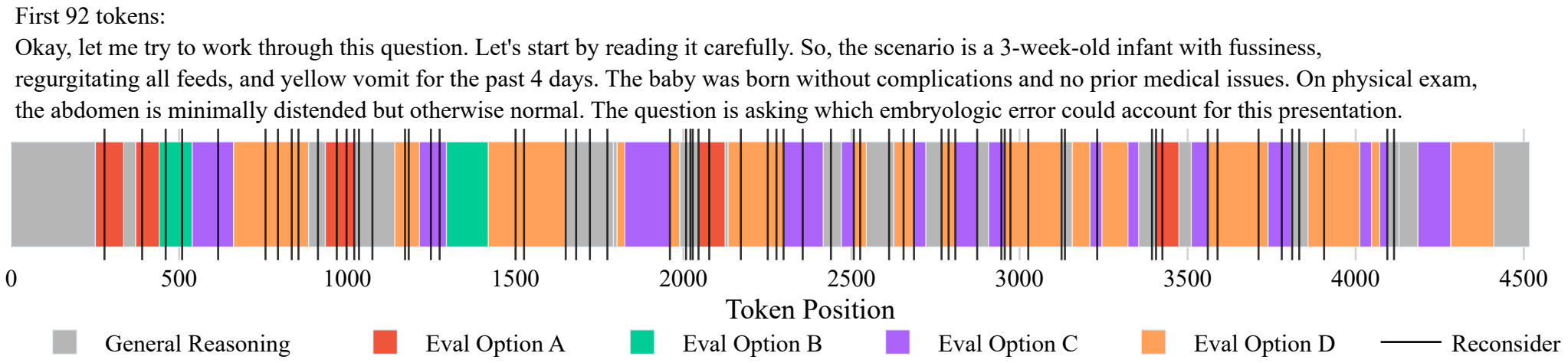}
\caption{Example reasoning data from M23k dataset. Among the 4517 tokens, the teacher model reason back and forth about different options, and pauses to reconsider its answer multiple times.}
\label{fig:reasoning-noisy}
\end{figure}
\section{Perturbation selection algorithms}
\label{app:perturbation-algorithms}

This section details the algorithm described in the main body. 

\paragraph{Inflection-point search (\Cref{alg:inflection-search}).}
Given a direction $\hat{u}$ and base parameters $\theta_0$, we approximate the curve $\epsilon \mapsto \ell(\epsilon)$ with a cubic spline fit to queried points and refine the bracket until the second derivative of the surrogate crosses zero. Cubic functions are especially suitable for this purpose: they are smooth and have a second derivative that is easy to compute, which enables us to find the inflection point efficiently by solving numerically for the root. Further, when perturbations are small, the third-order term will be small and the cubic function behaves like a parabola, which is consistent with the loss landscape behavior around the pretrained checkpoint; as we move away from the pretrained checkpoint, the third-order term still dominates other higher order terms, and captures the non-convex behavior of the loss landscape. In the algorithm we actively shrink or expand the search interval to ensure that the cubic function is still a good approximation of the loss curve.

\paragraph{LoRA train-and-perturb (\Cref{alg:lora-train-perturb}).}
To probe multiple axes without relying on a single random direction, we fit a rank-$r$ LoRA adapter, orthogonalize its rank-one factors $\{v_i\}$, and run \Cref{alg:inflection-search} along each $v_i$, as motivated by \Cref{thm:loss=grad} in the main body. \Cref{thm:loss=grad} also suggests that the perturbation scales should be far away from each other. We cap one perturbation scale to the inflection scale of the corresponding direction, as that is the maximum perturbation we are allowed to use to stay within the parabolic shape of the loss landscape; as for the other scale, we set it to the halfway point between the base loss and the inflection loss to maximize the richness of information in the loss signature. It ensures that the loss is different from the previous inflection loss, and also ensures that the perturbation are not too close to the pretrained checkpoint so that the loss of different directions are not too similar.

\paragraph{End-to-end pipeline (\Cref{alg:pipeline}).}
\Cref{alg:pipeline} stitches difficulty filtering, and diversity filtering together into a complete pipeline for data curation. $\Call{Generate}{\theta,D,k}$ refers to using the teacher model $\theta$ to (continue to) generate the response for the questions / partial responses in the dataset $D$ up to $k$ tokens.
Note that for perturbations, we only need to sample a small number of $N_{\mathrm{sample}}$ data samples to probe the loss landscape or finetune the LoRA model, which is extremely efficient. The perturbed checkpoints will evaluate all the problems that are not filtered out at that stage.

We follow the hyperparameter settings in \citet{ICML} to use $k_{\mathrm{difficulty}}=100$ and $k_{\mathrm{diversity}}=1000$.

\begin{algorithm}[t]
\caption{Inflection-point search along one direction}
\label{alg:inflection-search}
\begin{algorithmic}[1]
\Require Perturbation direction $\hat{u}$, student parameters $\theta_0$, dataset $\mathcal{D}$ of $(x,y)$ pairs
\Ensure Inflection scale $\epsilon^\star$
\State For $\epsilon\ge 0$, define $\ell(\epsilon)\gets\mathbb{E}_{(x,y)\sim\mathcal{D}}\!\left[\mathcal{L}_{problem}^k(x,y,\theta_0+\epsilon\hat{u})\right]$
\State Initialize coarse scales $\mathcal{E}\subset[0,\epsilon_{\max}]$
\State Initialize history $\mathcal{H}\gets\{(\epsilon,\ell(\epsilon)):\epsilon\in\mathcal{E}\}$
\For{$t=1,2,\ldots$}
    \State Reduce $\epsilon_{\max}$ if queried losses are not monotonically increasing \Comment{Evaluations are far away from the inflection point}
    \State Fit cubic spline surrogate $\hat{\ell}_t$ to $\mathcal{H}$
    \State $\epsilon_{\mathrm{cand}}\gets$ Root of $\frac{d^2}{d\epsilon^2}\hat{\ell}_t(\epsilon) = 0$
    \If{$\epsilon_{\mathrm{cand}}\notin[0,\epsilon_{\max}]$}
        \State Expand $\epsilon_{\max}$, query additional scales, and append to $\mathcal{H}$
        \State \textbf{continue}
    \EndIf
    \State Query $L_{\mathrm{cand}}\gets\ell(\epsilon_{\mathrm{cand}})$ and append $(\epsilon_{\mathrm{cand}},L_{\mathrm{cand}})$ to $\mathcal{H}$
    \State \textbf{if} $\epsilon^\star \approx \epsilon_{\mathrm{cand}}$ \textbf{then break} \textbf{else} $\epsilon^\star\gets\epsilon_{\mathrm{cand}}$ 
    \State \textbf{if} a fixed iteration cap is reached \textbf{then break} 
\EndFor
\State $\ell^\star\gets\ell(\epsilon^\star)$
\State \Return $(\epsilon^\star,\ell^\star, \hat{\ell}_t)$
\end{algorithmic}
\end{algorithm}

\begin{algorithm}[t]
\caption{LoRA train-and-perturb for directional checkpoints}
\label{alg:lora-train-perturb}
\begin{algorithmic}[1]
\Require Pretrained checkpoint $\theta_0$, dataset $D$, LoRA rank $r$
\Ensure Directional checkpoints $\{\theta_i^{(\mathrm{infl})},\theta_i^{(\mathrm{half})}\}_{i=1}^r$
\State Train a rank-$r$ LoRA adapter from $\theta_0$ on $D$
\State Decompose the adapter into orthogonal rank-1 directions $\{v_i\}_{i=1}^r$
\State $L_0\gets\mathbb{E}_{(x,y)\sim D}\!\left[\mathcal{L}_{problem}^k(x,y,\theta_0)\right]$
\For{$i=1,\dots,r$ \textbf{in parallel}}
    \State $(\epsilon_i^{(\mathrm{infl})},L_i^{(\mathrm{infl})}, \hat{\ell}_i)\gets$ Algorithm~\ref{alg:inflection-search}$(v_i,\theta_0,D)$
    \State $\epsilon_i^{(\mathrm{half})}\gets$ Root of $\hat{\ell}_i(\epsilon) = \left(L_0+L_i^{(\mathrm{infl})}\right)/2$
    \State $\theta_i^{(\mathrm{infl})}\gets\theta_0+\epsilon_i^{(\mathrm{infl})}v_i, \quad\theta_i^{(\mathrm{half})}\gets\theta_0+\epsilon_i^{(\mathrm{half})}v_i$
\EndFor
\State \Return $\{\theta_i^{(\mathrm{half})},\theta_i^{(\mathrm{infl})}\}_{i=1}^r$
\end{algorithmic}
\end{algorithm}

\begin{algorithm}[t]
\caption{Complete \method\  pipeline}
\label{alg:pipeline}
\begin{algorithmic}[1]
\Require Student model $\theta_0$, teacher model $\theta_{\mathrm{teacher}}$, seed question set $D$, target SFT size $N$
\Require Hyperparameters: Prefix length $k_{\mathrm{difficulty}}$ and $k_{\mathrm{diversity}}$ for teacher rollouts; Sample size $N_{\mathrm{sample}}$; LoRA rank $r$
\Ensure Curated dataset
\State $D_{\mathrm{difficulty}}\gets\Call{Generate}{\theta_{\mathrm{teacher}},D,k_{\mathrm{difficulty}}}$ 
\State Sample a random unit direction $\hat{u}$ in parameter space
\State $(\epsilon^\star,\cdot, \hat{\ell})\gets$ Algorithm~\ref{alg:inflection-search}$(\hat{u},\theta_0,\Call{Sample}{D_{\mathrm{difficulty}},N_{\mathrm{sample}}})$
\State $\theta_{\mathrm{diff}}\gets\theta_0+\epsilon^\star\cdot \hat{u}$
\State For each $(x_j,y_j)\in D_{\mathrm{difficulty}}$, compute $d_j\gets \ell^{k_{\mathrm{difficulty}}}(x_j,y_j,\theta_{\mathrm{diff}})$ 
\State Cluster $d_j$ by source into two groups; let $D_{\mathrm{hard}}$ be the high-loss cluster for each source
\State $D_{\mathrm{diversity}}\gets\Call{Generate}{\theta_{\mathrm{teacher}},D_{\mathrm{hard}},k_{\mathrm{diversity}}}$ 
\State $\bm{\Theta}_{\mathrm{pert}}\gets$ Algorithm~\ref{alg:lora-train-perturb}$(\theta_0,\Call{Sample}{D_{\mathrm{hard}},N_{\mathrm{sample}}},r)$ 
\State Build loss signature:  $\forall(x,y)\in D_{\mathrm{diversity}}$, form $L_j\gets\big(\ell^{k_{\mathrm{diversity}}}(x,y,\theta)\big)_{\theta\in\bm{\Theta}_{\mathrm{pert}}}$
\State Cluster $\{L_j\}$ (\Cref{eq:signature}) by source with number of clusters determined by \Cref{eq:budget}
\State $D_{\mathrm{sel}}\gets$ sample the same number of examples from each cluster
\State \Return $\Call{GenerateToCompletion}{\theta_{\mathrm{teacher}},D_{\mathrm{sel}}}$
\end{algorithmic}
\end{algorithm}

\section{Additional information and results on experiments}
\label{app:add-expr}
\subsection{Datasets and benchmarks}
\paragraph{Medical Reasoning} We follow the setting of \citet{M23k}, which curated a M1k dataset containing 1k examples from the M23k dataset, containing 4 sources form 23k examples. It comes with 10 evaluation benchmarks containing a total of 11k medical reasoning questions. The evaluation benchmark includes 3 in-distribution (ID) ones (MedMCQA \citep{pal2022medmcqa}, MedQA-USMLE \citep{jin2021disease}, and PubMedQA \citep{jin2019pubmedqa}), and 7 out-of-distribution (OOD) ones (MMLU-Pro (Medical) \citep{wang2024mmlu}, GPQA (Medical) \citep{rein2024gpqa}, Lancet and NEJM clinical case reports, MedBullets \citep{chen2024benchmarking}, MedXpertQA \citep{zuo2025medxpertqa}).

\paragraph{Math Reasoning} NuminaMath \citep{numinamath} dataset contains 859k math problems with ground truth answers. Among them, \citet{openthoughts} considered only 4 sources, namely ``amc\_aime'', ``aops\_forum'', ``math'', ``olympiads''. These 4 sources contain 192k data and they selected 89k of them and have Deepseek-R1 model to generate the reasoning traces. We use the same four sources and considerred QwQ-32B to generate the examples. For evaluation, the 4 benchmarks we use are the 4 math benchmarks presented in \citet{openthoughts}, namely MATH500 \citep{math500}, AMC23, AIME24, AIME25 \citep{maa_aime_2024}. The latter three benchmarks are relatively small, and we repeat the evaluation 6 times for a more accurate evaluation of mean and std. 

\paragraph{Physics reasoning}  We combine two datasets for training, namely \citet{physicsEval} containing 17,635 textbook physics questions (12 removed for empty questions or ground-truth answers) and 3084 olympiad questions we curated by ourselves. The olympiad dataset originates from \citet{physicsolympicoriginate}, yet their dataset contains OCR errors, subquestions, and images. We prompted LLMs with few-shot prompt to sanitize the errors, split the subquestions while carrying on all necessary information from presious subquestions, and describe the image or discard the problem depending on the nature of the image. We end up repeating harder but smaller subsets to ensure that we have enough questions to select from these harder questions. Specifically, we repeat the questions from WoPho, RMPh, EuPhO 16 times, IPhO 4 times, NbPhO 9 times, USAPhO 3 times, and APhO 6 times. These numbers are selected based on the desired number of data to select from according to our algorithm. Combined with the 17,635 textbook questions, we end up with 39595 physics questions to curate the SFT dataset from.
For evaluation, We use the same evaluation benchmark as \citet{p1-2025}, except that we only evaluated on the text data as the scope of this work is text reasoning. The benchmark contains 102 questions from seven 2024–2025 competitions, and we repeat the evaluation 3 times to take the average.

\subsection{Data Decontamination} We follow the same decontamination protocol as the one used in \citet{M23k} and \citet{openthoughts}, where a fuzzy search is applied. Specifically, the normalized indel similarity is calculated, which is based on how much insertions and deletion is required to turn an example from the train set into one in the test set. This identifies not only exact matches, but also similar questions with minor variations. We decontaminated ourselves for the numina dataset; we verified that the Physics dataset is not contaminated, which is expected because our training set only incluses textbook problems from \citet{physicsEval} and Olympic questions we curated ourselves from 1967-2018. We also verified that the M23k dataset provided by \citet{M23k} is already decontaminated as claimed by the authors.

\subsection{Hyperparameters} We randomly sample a subset of data containing 800 examples for training with LoRA to obtain the directions. We use LoRA rank 4 for all tasks other than math reasoning (which we use LoRA rank 2 as it's more efficient). We train LoRA for 200 iterations. This is especially efficient: For instance, in the setting of data generation for medical with QwQ-32B, the entire cost for generating LoRA examples and training LoRA is only 16 minutes.
Other hyperparameters are consistent with the ones in \cite{ICML}. Specifically, the temperature was set to 0.7 for math and physics and 0 for medical reasoning, and all other generation configs were engine defaults. Since Qwen2.5 models have a max context of 32k tokens, we cap response length to 27k tokens for math and physics to set aside 5k tokens for prompt. For medical, we cap the max response length to 16k tokens to match the implementation in \citet{M23k}. Without specification, we repeat the inference on each benchmark twice to get the mean and std.

\subsection{Baselines} Baselines in medical reasoning and math reasoning are consistent with \citet{ICML}. This includes:
\begin{itemize}
    \item Learnability \citep{DavIR-learnability-loss}: Learnability measures how much loss of an example decreases during the training and selects examples that are most learnable by the model, i.e. experience the largest drop in loss during training.
    \item Middle Perplexity \citep{marion2023less}: Given a trained model, it ranks all the data by perplexity and takes the required number of data around the median. The assumption is that easiest examples with lowest perplexity do not benefit training and hardest examples with highest perplexity are noisy or outliers.
    \item Embedding Clustering \citep{bhatt2024experimental}: Given a decoder-only LLM, the method takes the last hidden state as the feature, and uses k-center algorithm on the embedding to pick a set of diverse examples.
    \item TEMP \citep{ICML}: This method extends \citet{S2L} to long reasoning datasets by identifying the initial tokens reflects data quality. It fine-tunes the model on the full SFT data and tracks the loss values of the initial reasoning tokens for all examples at noisy checkpoints along the fine-tuning directions. Then, it clusters the loss trajectories and samples examples with higher loss values from the clusters. This approach fundamentally depends on access to the full SFT dataset and thus cannot be applied for data generation.

\end{itemize}

\section{Additional details and discussions on teacher selection}
\label{app:teacher}
In this section, we provide additional details and results on teacher selection. 

\paragraph{Loss evaluation} We detail how the blue bars in \Cref{fig:teacher-selection}  are calculated. We randomly select 1000 questions from the NuminaMath dataset and prompt the teacher model to generate the reasoning traces with up to $150$ tokens, with the extra 50 tokens to account for the variation in tokenization, using the recommended generation parameters for each model individually. For teacher selection, we prompt rach teacher model to ``Explain what the problem is asking in your own words.'' The pretrained student model evaluates the first 100 tokens of the teachers generation using the standard answering prompt, namely, ``\texttt{Return your final response within \textbackslash boxed\{\}. \{question\}}'' as if the teacher model started its reasoning with rephrasing the problem.

\paragraph{Finetuning the student} We detail how the green bars in \Cref{fig:teacher-selection} are calculated. We randomly select 1000 questions from the NuminaMath dataset and generate with teacher model up to 16384 tokens with the prompt ``\texttt{Return your final response within \textbackslash boxed\{\}. \{question\}}'' using the recommended generation parameters for each model individually to form a SFT dataset for each teacher model. We then finetune the student model (Qwen2.5-7B-Instruct) on the generated dataset and evaluate the accuracy on the test set. The performance is shown in \Cref{tab:teacher-selection}.

\begin{table}[ht]
    \centering
    \caption{Average accuracy (\%) of the student model finetuned on the a same set of 1000 randomly selected questions from the NuminaMath dataset, generated by different teacher models.}
    \label{tab:teacher-selection}
    \begin{footnotesize}
    \begin{tabular}{l c c c c c}
    \toprule
    Teacher & Avg. & AIME24 & AIME25 & AMC23 & MATH500 \\
    \midrule
    Qwen/Qwen3.6-27B & $14.74 \pm 0.62$ & $0.00 \pm 0.00$ & $0.00 \pm 0.00$ & $16.88 \pm 2.39$ & $42.10 \pm 0.14$ \\
    Openai/GPT-OSS-120B & $14.52 \pm 0.82$ & $2.78 \pm 2.51$ & $1.11 \pm 1.72$ & $19.58 \pm 2.46$ & $34.60 \pm 0.00$ \\
    Google/Gemma-4-26B-A4B-IT & $24.88 \pm 0.35$ & $7.78 \pm 1.72$ & $3.33 \pm 0.00$ & $32.50 \pm 0.00$ & $55.90 \pm 0.71$ \\
    Google/Gemma-4-31B-IT & $34.43 \pm 0.60$ & $8.33 \pm 1.92$ & $13.33 \pm 2.72$ & $46.25 \pm 1.44$ & $69.80 \pm 0.00$ \\
    Microsoft/Phi-4-Reasoning-Plus & $37.26 \pm 2.94$ & $19.44 \pm 8.28$ & $13.33 \pm 0.00$ & $46.67 \pm 3.76$ & $69.60 \pm 0.00$ \\
    Qwen/QwQ-32B & $43.56 \pm 1.23$ & $15.83 \pm 1.67$ & $18.33 \pm 5.77$ & $61.88 \pm 1.25$ & $78.20 \pm 0.28$ \\
    \bottomrule
    \end{tabular}
    \end{footnotesize}
\end{table}

\paragraph{Discussion on the necessity of teacher rephrasing} 
In \Cref{eq:teacher-rephrasing}, we rank the teachers based on the rephrasing loss instead of the the loss of the actual reasoning in the solution given by the teacher. 
This is because selection based on rephrasing and reasoning optimize for opposing dynamics, and applying the same low-loss selection strategy to the first part of a reasoning chain fundamentally degrades performance. While prompt rephrasing merely changes the problem’s external framing, the initial steps of a reasoning trace determine its entire trajectory. Selecting reasoning steps based on low student loss systematically discards the teacher’s most valuable contributions--such as novel abstraction pivots, non-obvious problem decompositions, and counter-intuitive intermediate steps-—simply because they are unfamiliar to the student. Instead of encouraging the student to learn new, sophisticated problem-solving pathways, this filtering mechanism forces the model into a local minimum, prioritizing predictability over logical correctness.

While most models start their response with summarizing or rephrasing the problem, there are counterexamples like Gemma-4 series and Qwen3.6 series models, which directly start solving the problem. The rephrasing prompt we use therefore ensures that we're measuring the same metric across different teacher models, namely how good can the student follow the teacher's rephrasing of the problem itself.
Our additional ablation on teacher selection shown in \Cref{tab:teacher-mean-loss-no-rephrase} confirms that, without prompting the teacher to explain the problem, the loss of the initial reasoning tokens does not correlate with finetune accuracy:
\begin{table}[ht]
    \centering
    \caption{Loss of initial reasoning tokens vs. the average finetune accuracy reported in \Cref{fig:teacher-selection}. Without prompting the teacher to explain the problem, the loss of the initial reasoning tokens does not correlate with finetune accuracy.}
    \label{tab:teacher-mean-loss-no-rephrase}
    \begin{tabular}{l c c}
    \toprule
    Teacher & Mean Loss & Avg. Finetune Acc. \\
    \midrule
    Google/Gemma-4-26B-A4B-IT & $1.0045 \pm 0.311$ & $24.88\% \pm 0.35\%$ \\
    Google/Gemma-4-31B-IT & $1.0049 \pm 0.295$ & $34.43\% \pm 0.60\%$ \\
    Qwen/QwQ-32B & $1.3022 \pm 0.302$ & $43.56\% \pm 1.23\%$ \\
    Qwen/Qwen3.6-27B & $1.3467 \pm 0.415$ & $14.74\% \pm 0.62\%$ \\
    Openai/GPT-OSS-120B & $1.7742 \pm 0.445$ & $14.52\% \pm 0.82\%$ \\
    Microsoft/Phi-4-Reasoning-Plus & $1.8296 \pm 0.479$ & $37.26\% \pm 2.94\%$ \\
    \bottomrule
    \end{tabular}
\end{table}
\section{Empirical evidence of the increase of diversity}
\label{app:emp-div}
\begin{figure}
    \centering
    \includesvg[width=0.8\linewidth]{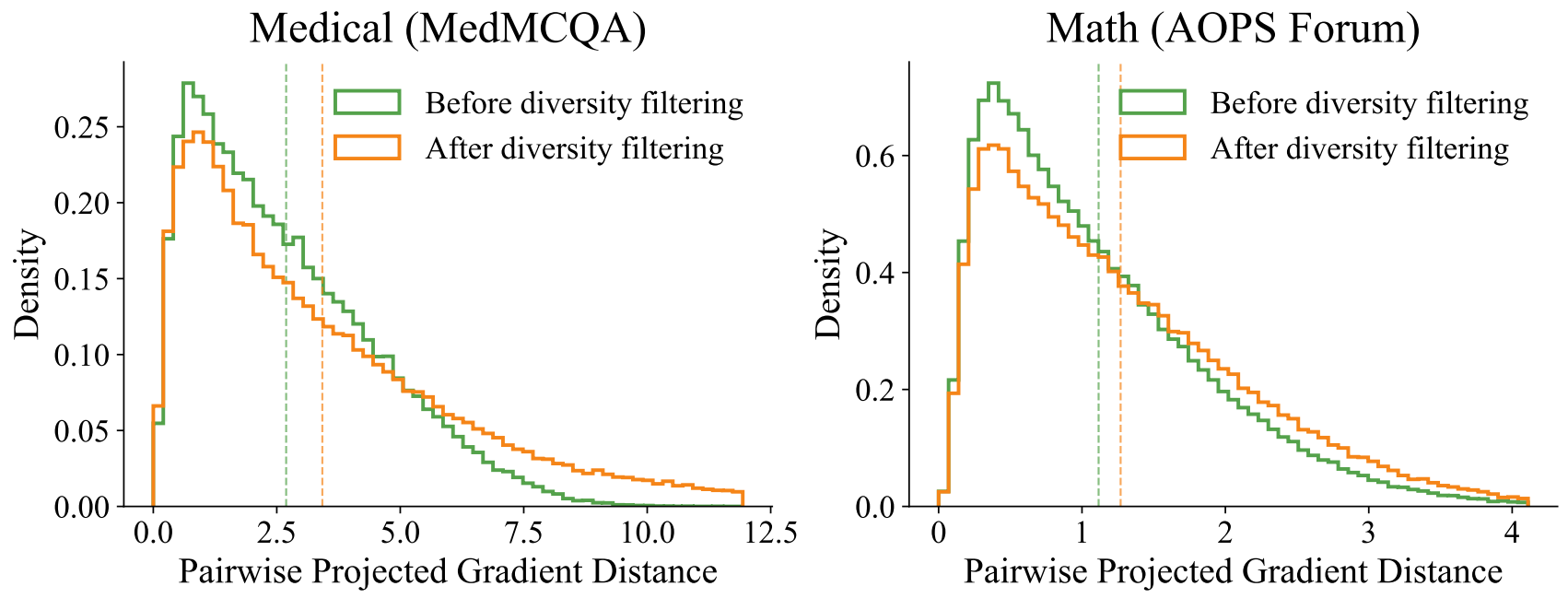}
    \caption{The distribution of the pairwise projected gradient difference before and after diversity filtering. Vertical line measures the mean. Diversity filtering increased the pairwise difference, indicating an increases in the diversity of the gradients.}
    \label{fig:fig:emp-div}
\end{figure}

The diversity metrics we aim to optimize for is the gradient diversity along the LoRA directions. To quantify the increment in diversity, we calculate the norm of pairwise gradient difference projected on the LoRA dimensions. Specifically, we take all the data after diversity filtering of a source, and take the same number of data from the same source before diversity filtering. With the full gradients at the pretrained checkpoint $g^i$ for example $i$, and the LoRA directiosn being \[V=\begin{pmatrix}
    v_1^\top\\
    v_2^\top \\
    \vdots\\
    v_r^\top
\end{pmatrix}\] where $r$ is the LoRA rank, we calculate the distribution of \[\Delta_{ij}=\left\|Vg_i-Vg_j\right\|_2\] and present in \Cref{fig:fig:emp-div}.

The root-mean-square of the norm, namely $\sqrt{\E_{i\neq j}\Delta_{ij}^2}$, changes from 3.26 to 4.41 (35.2\% increase) and 1.36 
to 1.55 (13.8\% increase) on medical and math respectively, verifying the increased diversity.
\newpage
\section*{NeurIPS Paper Checklist}

The checklist is designed to encourage best practices for responsible machine learning research, addressing issues of reproducibility, transparency, research ethics, and societal impact. Do not remove the checklist: {\bf The papers not including the checklist will be desk rejected.} The checklist should follow the references and follow the (optional) supplemental material.  The checklist does NOT count towards the page
limit. 

Please read the checklist guidelines carefully for information on how to answer these questions. For each question in the checklist:
\begin{itemize}
    \item You should answer \answerYes{}, \answerNo{}, or \answerNA{}.
    \item \answerNA{} means either that the question is Not Applicable for that particular paper or the relevant information is Not Available.
    \item Please provide a short (1--2 sentence) justification right after your answer (even for \answerNA). 
\end{itemize}

{\bf The checklist answers are an integral part of your paper submission.} They are visible to the reviewers, area chairs, senior area chairs, and ethics reviewers. You will also be asked to include it (after eventual revisions) with the final version of your paper, and its final version will be published with the paper.

The reviewers of your paper will be asked to use the checklist as one of the factors in their evaluation. While \answerYes{} is generally preferable to \answerNo{}, it is perfectly acceptable to answer \answerNo{} provided a proper justification is given (e.g., error bars are not reported because it would be too computationally expensive'' or ``we were unable to find the license for the dataset we used''). In general, answering \answerNo{} or \answerNA{} is not grounds for rejection. While the questions are phrased in a binary way, we acknowledge that the true answer is often more nuanced, so please just use your best judgment and write a justification to elaborate. All supporting evidence can appear either in the main paper or the supplemental material, provided in appendix. If you answer \answerYes{} to a question, in the justification please point to the section(s) where related material for the question can be found.



\begin{enumerate}

\item {\bf Claims}
    \item[] Question: Do the main claims made in the abstract and introduction accurately reflect the paper's contributions and scope?
    \item[] Answer: \answerYes{} 
    \item[] Justification: All claims have been accurately reflected.

\item {\bf Limitations}
    \item[] Question: Does the paper discuss the limitations of the work performed by the authors?
    \item[] Answer: \answerYes{} 
    \item[] Justification: We acknowledge the requirement of minimal training and not curating a larger dataset.

\item {\bf Theory assumptions and proofs}
    \item[] Question: For each theoretical result, does the paper provide the full set of assumptions and a complete (and correct) proof?
    \item[] Answer: \answerYes{} 
    \item[] Justification: Complete proof is given in Appendix A.

    \item {\bf Experimental result reproducibility}
    \item[] Question: Does the paper fully disclose all the information needed to reproduce the main experimental results of the paper to the extent that it affects the main claims and/or conclusions of the paper (regardless of whether the code and data are provided or not)?
    \item[] Answer: \answerYes{} 
    \item[] Justification: Detailed hyperparameters have been provided.

\item {\bf Open access to data and code}
    \item[] Question: Does the paper provide open access to the data and code, with sufficient instructions to faithfully reproduce the main experimental results, as described in supplemental material?
    \item[] Answer: \answerYes{} 
    \item[] Justification: The code and data is provided in the link in the appendix, which contains reproduction instructions.

\item {\bf Experimental setting/details}
    \item[] Question: Does the paper specify all the training and test details (e.g., data splits, hyperparameters, how they were chosen, type of optimizer) necessary to understand the results?
    \item[] Answer: \answerYes{} 
    \item[] Justification: We include all experimental details in the appendix

\item {\bf Experiment statistical significance}
    \item[] Question: Does the paper report error bars suitably and correctly defined or other appropriate information about the statistical significance of the experiments?
    \item[] Answer: \answerYes{} 
    \item[] Justification: We included error bars and stds in our plots.

\item {\bf Experiments compute resources}
    \item[] Question: For each experiment, does the paper provide sufficient information on the computer resources (type of compute workers, memory, time of execution) needed to reproduce the experiments?
    \item[] Answer: \answerYes{} 
    \item[] Justification: Details are included in the appendix.
    
\item {\bf Code of ethics}
    \item[] Question: Does the research conducted in the paper conform, in every respect, with the NeurIPS Code of Ethics \url{https://neurips.cc/public/EthicsGuidelines}?
    \item[] Answer: \answerYes{} 
    \item[] Justification: We conform the Code of Ethics strictly.

\item {\bf Broader impacts}
    \item[] Question: Does the paper discuss both potential positive societal impacts and negative societal impacts of the work performed?
    \item[] Answer: \answerYes{} 
    \item[] Justification: We do not have negative impact. Our method facilitates efficient AI and is environmentally friendly.
    
\item {\bf Safeguards}
    \item[] Question: Does the paper describe safeguards that have been put in place for responsible release of data or models that have a high risk for misuse (e.g., pre-trained language models, image generators, or scraped datasets)?
    \item[] Answer: \answerNA{} 
    \item[] Justification: We do not pose such risk.

\item {\bf Licenses for existing assets}
    \item[] Question: Are the creators or original owners of assets (e.g., code, data, models), used in the paper, properly credited and are the license and terms of use explicitly mentioned and properly respected?
    \item[] Answer: \answerYes{} 
    \item[] Justification: We have cited all original works and respected the terms.

\item {\bf New assets}
    \item[] Question: Are new assets introduced in the paper well documented and is the documentation provided alongside the assets?
    \item[] Answer: \answerYes{} 
    \item[] Justification: We have released our code and dataset assets with proper documentation.

\item {\bf Crowdsourcing and research with human subjects}
    \item[] Question: For crowdsourcing experiments and research with human subjects, does the paper include the full text of instructions given to participants and screenshots, if applicable, as well as details about compensation (if any)? 
    \item[] Answer: \answerNA{} 
    \item[] Justification: We do not involve crowdsource.

\item {\bf Institutional review board (IRB) approvals or equivalent for research with human subjects}
    \item[] Question: Does the paper describe potential risks incurred by study participants, whether such risks were disclosed to the subjects, and whether Institutional Review Board (IRB) approvals (or an equivalent approval/review based on the requirements of your country or institution) were obtained?
    \item[] Answer: \answerNA{} 
    \item[] Justification: We do not involve crowdsource.

\item {\bf Declaration of LLM usage}
    \item[] Question: Does the paper describe the usage of LLMs if it is an important, original, or non-standard component of the core methods in this research? Note that if the LLM is used only for writing, editing, or formatting purposes and does \emph{not} impact the core methodology, scientific rigor, or originality of the research, declaration is not required.
    \item[] Answer: \answerYes{} 
    \item[] Justification: We followed the LLM policy. 

\end{enumerate}
\end{document}